\documentclass[compsoc,onecolumn]{IEEEtran}
\usepackage[T1]{fontenc}
\usepackage[utf8]{inputenc}
\usepackage{babel}
\usepackage{array}
\usepackage{float}
\usepackage{mathtools}
\usepackage{amsmath}
\usepackage{amssymb}
\usepackage{graphicx}
\usepackage{xcolor}
\usepackage[unicode=true]{hyperref}
\usepackage{authblk}

\usepackage[authoryear,longnamesfirst]{natbib}
\makeatletter

\providecommand{\tabularnewline}{\\}
\floatstyle{ruled}
\newfloat{algorithm}{tbp}{loa}
\providecommand{\algorithmname}{Algorithm}
\floatname{algorithm}{\protect\algorithmname}

\let\oldforeign@language\foreign@language
\DeclareRobustCommand{\foreign@language}[1]{%
  \lowercase{\oldforeign@language{#1}}}
  
\usepackage{algorithmic}
\usepackage{tikz}
\usetikzlibrary{bayesnet}
\usetikzlibrary{shadings}
\usepackage{caption}
\usepackage{subcaption}
\usetikzlibrary{positioning}

\makeatother

\begin{document}
\let\WriteBookmarks\relax
\def\floatpagepagefraction{1}
\def\textpagefraction{.001}

\title{Fully Bayesian Approaches to Topics over Time}

\author[1,2,*]{Julián Cendrero \thanks{* Corresponding author: Julián Cendrero (jcendrero3@alumno.uned.es).\newline Work completed while the author was at mrHouston Tech Solutions and UNED.}}
\author[1]{Julio Gonzalo}
\author[3]{Ivar Zapata}
\affil[1]{Universidad Nacional de Educación a Distancia (UNED)}
\affil[2]{Cohere}
\affil[3]{mrHouston Tech Solutions \vspace{1.5ex}} 

\renewcommand{\Authfont}{\bfseries}
\renewcommand{\Affilfont}{\mdseries \itshape}

\IEEEtitleabstractindextext{
\begin{abstract}
The Topics over Time (ToT) model captures thematic changes in timestamped datasets by
explicitly modeling publication dates jointly with word co-occurrence patterns.
However, ToT was not approached in a fully Bayesian fashion, a flaw that
makes it susceptible to stability problems. To address this issue, we propose a fully
Bayesian Topics over Time (BToT) model via the introduction of a conjugate
prior to the Beta distribution. This prior acts as a regularization
that prevents the online version of the algorithm from unstable updates
when a topic is poorly represented in a mini-batch. The characteristics
of this prior to the Beta distribution are studied here for the first
time. Still, this model suffers from a difference in scale between the single-time observations and the multiplicity of words per document. A variation of BToT, Weighted Bayesian Topics over Time (WBToT), is proposed as a solution. In WBToT, publication dates are repeated a certain
number of times per document, which balances the relative influence
of words and timestamps along the inference process. We have tested
our models on two datasets: a collection of over 200 years of US state-of-the-union (SOTU) addresses and a large-scale COVID-19 Twitter corpus of 10 million tweets. The results show
that WBToT captures events better than Latent Dirichlet Allocation and other SOTA topic models like BERTopic: the median absolute deviation of the topic presence over time is reduced by $51\%$ and $34\%$, respectively. Our experiments also demonstrate the superior coherence of WBToT over BToT, which highlights the importance of balancing the time and word modalities. Finally, we illustrate the stability of the online optimization algorithm in WBToT, which allows the application of WBToT to problems that are intractable for standard ToT.
\end{abstract}

}

\maketitle


\section{\label{sec:Introduction}Introduction}

Topic modeling is widely acknowledged as a method for discovering
the latent themes that characterize a collection of documents (for
recent reviews with emphasis on applications, see \citealt{Churchill2021, Jelodar2019}).
Mathematically, a topic is defined as a multinomial distribution over
a collection of words that constitute a vocabulary. A document can be modeled as a multinomial
distribution over topics, each of them representing the underlying
semantic themes of the document. The seminal work on topic modeling,
Probabilistic Latent Semantic Analysis (PLSA) \citep{Hofmann1999},
was later extended in \citet{Blei2003,Hoffman2010} by making use of Bayesian methods
to assemble the predominant Latent Dirichlet Allocation
(LDA). The shortcomings of PLSA (namely,
a large number of parameters and difficulties to handle new documents
unseen during training) were solved in LDA thanks to the introduction of conjugate
priors for each probability distribution.

LDA was the origin of an abundance of related models which extend its
capabilities by including additional features like word embeddings \citep{Bunk2018, Dieng2020} and hierarchical structure \citep{Viegas2020}, or by adapting the generative process to fit specific types of documents (such as short texts, \citealt{Li2019a}).

Publication dates have been leveraged to investigate topic evolution in timestamped
corpora. Modeling time-related characteristics of texts has been proven to be effective for many purposes, such as disaster detection, public opinion evaluation, and outbreak prediction \citep{Asghari2020}. There are two main strategies to take publication timestamps into account in topic modeling. First, one can see the topics as dynamic objects. This way, the documents in a given time slice are modeled by a set of topics
that evolved from the documents of an adjacent time slice. This
is the idea behind the Dynamic Topic Model (DTM) \citep{Blei2006,Wang2008}.
On the other hand, one can assume that topics are static (like
in LDA) and associate each topic with a continuous distribution over
timestamps. This approach is the basis for the Topics over Time (ToT)
model \citep{Wang2006}, whose generative process models documents
and timestamps jointly. The main advantage of such a continuous time
model over DTM is the fact that there is no need to arbitrarily define
the size of a discrete-time slice.

The ToT model was designed for collections of timestamped documents
that were published over a large period of time. In this type of
corpus, topics that appear at the beginning of the period can be quite
different from the ones that appear later: new trends emerge and others
disappear through time. A time-unaware model such as LDA will extract topics that are present across the
entire corpus. This is where the ToT model delivers an advantage:
since this model associates each topic with a continuous distribution
over timestamps, the projection of a document into the topic space
will not only take into account word co-occurrences but also the
document's timestamp. As a result, ToT will partition topics that are lexically close but show different time distributions.

Some recent applications of the ToT model include the detection
of public concerns in mega infrastructure projects \citep{Xue2020,Xue2021},
time series analysis of supercomputer logs to predict node failures
in high performance computing systems \citep{Das2018}, and meeting
summarization tools \citep{Shi2018}. For us, such a variety of applications
justify the interest in the ToT model and its potential improvements.

The main shortcoming of ToT is the fact that, unlike LDA, it is not
a fully Bayesian model. Apart from the theoretical inconvenience of
dealing with two different frameworks at the same time (a Bayesian
approach for the text modality and a frequentist approach for the time modality),
this results in stability problems. In particular, parameter
estimation in an online fashion is susceptible to unstable updates when
a topic is poorly represented in a mini-batch. The lack of a robust
online optimization method makes this model unusable for many practical
applications that require an online data stream mining (for example,
content-based recommendation systems \citealt{Al-Ghossein2018}). 
This limitation supports the need for a fully Bayesian approach to Topics over Time.
In this work, we solve this problem by introducing
a conjugate prior to the timestamp probability distribution. We call this model 
Bayesian ToT (BToT).

However, as we will later show,
we have found that this naive BToT suffers from a difference in scale between
word and timestamp observations in a single document.
In order to address this issue, we also introduce an extension of BToT, called Weighted Bayesian Topics over Time (WBToT), that dynamically
lowers the number of time observations. This mechanism is similar to the
introduction of a balancing hyperparameter in \citet{Wang2006,Fang2017}.
The WBToT model features a stable online optimization algorithm 
that makes it suitable for many applications where standard ToT is not applicable. 
The Bayesian formalism also simplifies many existing extensions of ToT \citep{Poumay2021}.

The structure of this paper is as follows. In Section \ref{sec:Related-work},
we will dive into the technical details of the ToT model and its extensions
to motivate our Bayesian approach. In Section \ref{sec:Contributions},
we specify the contributions of our work. In Section \ref{sec:models},
we introduce two original models, BToT and WBToT, and propose a variational
approach to optimization (both in batch and online fashions). In Section
\ref{sec:datasets} we describe the two datasets that will be employed
to validate our models and the experiments that will be performed. In Section \ref{sec:experimental-results} we present the main results
of our experiments and discuss their implications. We conclude with
some final remarks in Section \ref{sec:Conclusions}.

\section{Literature Review\label{sec:Related-work}}
The use of Bayesian methods in topic modeling was introduced via the
LDA model in \citet{Blei2003, Hoffman2010}. For each of the multinomial
distributions that constituted the original PLSA model, a conjugate
prior (Dirichlet distribution) was added. Although the resulting posterior distribution was intractable
for exact inference, an approximate inference was performed through
variational inference. A family of lower bounds indexed
by a set of free variational parameters is introduced, and its optimization could be
easily achieved via an iterative fixed-point method. For reference,
a detailed description of the LDA generative process and its optimization
algorithms (batch and online) is presented in Appendix \ref{sec:Description-of-LDA}.
We will follow the same approach and will borrow most of its notation
in order to turn ToT into a fully Bayesian model. 

In the ToT model \citep{Wang2006},
topics are created by looking not only at word co-occurrences but
also at the documents' timestamps. In the generative process for ToT,
a distribution of timestamps is postulated to come from a mixture of Beta distributions,
one per topic, with the same mixture parameters per document
as the one that generates the words. However, this model is not fully
Bayesian because there is no prior for the parameters of the Beta distributions,
so they have to be optimized by ordinary maximum likelihood. Gibbs sampling
is used as the optimization algorithm for this model, and an approximation
based on the method of moments simplifies the mathematics of the maximization. 

Gibbs sampling is not the only possible way for ToT parameter inference.
In \citet{Fang2017}, optimization of the ToT log-likelihood was made
via a variational ansatz which is identical to the one for LDA in
\citet{Hoffman2010}, together with an exact maximization of the evidence
lower bound (ELBO) with respect to both variational and Beta distribution
parameters. Note that, in this approach, the Beta distribution parameters
are maximized directly, with no prior distribution involved. Therefore,
although a variational inference approach is used instead of Gibbs
sampling for the optimization, the ToT model is not actually modified. Notice that neither of the two works \citep{Wang2006, Fang2017} introduced an online optimization approach for the model. For reference, a detailed description of the ToT generative process and its optimization algorithm is presented in Appendix \ref{sec:Description-of-ToT}.

Most extensions of ToT have focused on enhancing its capabilities by including
additional contextual features, but none of them modified the basic generative
process for the timestamps. For example, the author-topic over time
model (AToT) \citep{Xu2014} includes the document's authorship into its
generative model by the introduction of a uniform distribution over
authors. This way, they are able to model how the authors' interests
change over time, which has been successfully applied to real case
scenarios like monitoring emerging technologies by using Twitter data
mining \citep{Li2019}. However, they still use an unregularized distribution
for the timestamps, thus presenting the same problem as the original
ToT in terms of stability. 

The most recent extensions of ToT still suffer from this problem: the Hierarchical Topic Modeling Over Time (HTMOT) model
\citep{Poumay2021}, which models topic temporality and hierarchy jointly,
cannot be trained using variational inference due to the lack of a
conjugate prior for the beta distribution. Due to the prohibitively
long training time that standard Gibbs sampling takes for this model,
this shortcoming forced the authors to develop a new implementation
of Gibbs sampling. We believe that they could have employed a more
straightforward optimization algorithm if the original ToT model were Bayesian to begin with.

Unlike these extensions, our work is an attempt to modify the core generative of ToT process to make it fully Bayesian. We will start from the basic generative process introduced in \citet{Wang2006} but, instead of using Gibbs sampling, we will consider the variational inference method described by \citet{Fang2017}. We draw inspiration from \citet{Blei2003} and will introduce conjugate priors for each probability distribution. For the online version of the algorithm, we will closely resemble the approach introduced in \citet{Hoffman2010}.

It must be noticed that our work is based on probabilistic topic models. Recently, alternative approaches that make use of large pre-trained language models \citep{Vaswani2017, Devlin2019} have also been proposed. The most popular framework, BERTopic \citep{Grootendorst2022}, projects documents into an embedding space by making use of Sentence-BERT \citep{Reimers2019}, then performs dimensionality reduction through UMAP \citep{McInnes2018} and finally creates clusters with HDBSCAN \citep{McInnes2017}. Topic distributions are generated by using TF-IDF over such clusters. This approach reportedly yields more coherent topics than models based on the Bag of Words (BOW) assumption, but it does not leverage publication timestamps for event detection. Nevertheless, we will also include BERTopic in our experiments as an additional baseline.

\section{Research Objectives and Contributions\label{sec:Contributions}}

The goal of the present paper is to extend the model introduced in \citet{Wang2006}
and its subsequent variational approach in \citet{Fang2017} to make them fully Bayesian. In particular, our main contributions are:
\begin{enumerate}
\item Reformulation of ToT as a Bayesian model. This implies that
a novel distribution (which we will call Beta-prior \footnote{Here, we call Beta-prior to the conjugate prior to the Beta distribution.
Other works refer to Beta-prior as the Beta distribution seen as a
prior to a multinomial distribution, to distinguish it from the resulting
posterior, which is also a Beta distribution.}) has to be used as a prior for the parameters of the timestamp Beta
distributions. We call this model Bayesian Topics over Time (BToT).
\item Introduction of a modified version of BToT which dynamically
lowers the number of time observations. This mimics the introduction of a balancing hyperparameter in \citet{Wang2006,Fang2017},
which balances the relative influence of the two modalities (words
and timestamps) in the log-likelihood. We call this model Weighted
Bayesian Topics over Time (WBToT).
\item Calculation of a variational approach inference of parameters for
both BToT and WBToT. In both cases, we introduce a batch and an online
version.
\item Experimental validation on two datasets: a collection of over 200
years of US state-of-the-union addresses and a large-scale COVID-19
Twitter corpus.
\end{enumerate}

\section{Fully Bayesian Approaches to Topics over Time\label{sec:models}}

In this section, we present two novel models and propose variational
inference algorithms for parameter optimization: in Section \ref{subsec:BToT}
we introduce the naive Bayesian Topics over Time (BToT) model and
in Section \ref{subsec:WBToT} we introduce the Weighted Bayesian
Topics over Time (WBToT). For a summary of the notation employed in this section, see Table \ref{tbl:Notation_Observed} - \ref{tbl:NotationParameters}.

\subsection{Bayesian Topics over Time (BToT)\label{subsec:BToT}}

\subsubsection{Description of the model\label{subsubsec:btot-description}}

The ToT model introduced a Beta distribution of the form
\begin{equation}
p\left(t\left|\boldsymbol{\rho}_{k}\right.\right)=B\left(\boldsymbol{\rho}_{k}\right)^{-1}t^{\rho_{k}^{1}-1}\left(1-t\right)^{\rho_{k}^{2}-1}
\end{equation}
for the timestamp observation $t$ in topic $k$, where $B$ is the Beta function (employed here as a normalization constant). Notice that we have
denoted $\boldsymbol{\rho}_{k}=\left(\rho_{k}^{1},\rho_{k}^{2}\right)$
the 2D natural parameter of the Beta distribution, and hence $B\left(\boldsymbol{\rho}_{k}\right)=B\left(\rho_{k}^{1},\rho_{k}^{2}\right)$.
These two parameters must satisfy $\rho_{k}^{1}>0$ and $\rho_{k}^{2}>0$
to ensure that the distribution can be normalized. We will extend
the ToT model by adding a prior for this distribution. To the best
of our knowledge, no conjugate prior to the Beta distribution has
been studied so far. However, since the Beta distribution is a member
of the exponential family, one can always find a general conjugate
prior (see \citep{Bishop2007}) of the form
\begin{equation}
p\left(\boldsymbol{\rho}_{k}\left|\nu,\boldsymbol{\chi}\right.\right)=f\left(\nu,\boldsymbol{\chi}\right)\exp\left[\nu\left(\boldsymbol{\rho}_{k}\cdot\boldsymbol{\chi}-\log B\left(\boldsymbol{\rho}_{k}\right)\right)\right],\label{eq:beta-prior}
\end{equation}
where $f\left(\nu,\boldsymbol{\chi}\right)$ is a normalization function. We will call the distribution in Eq. (\ref{eq:beta-prior}) Beta-prior.
Note that, for simplicity of presentation, we use the same prior hyper-parameters
for all topics, but the formulas could be generalized straightforwardly
to include topic-dependent hyper-parameters. Since $\rho^{i}>0$,
integrability demands that $\nu>0$. The $\boldsymbol{\chi}$ parameters
are constrained by $\exp\chi^{1}+\exp\chi^{2}<1$, which can be proved
by transforming the normalization integral to polar coordinates, using
the Stirling approximation in $\log B\left(\boldsymbol{\rho}\right)$
for $\rho\rightarrow\infty$ and then demanding that the exponential
tends to zero at its largest asymptotic values (see Appendix \ref{sec:integrability-analysis} for a detailed explanation).

\begin{figure}
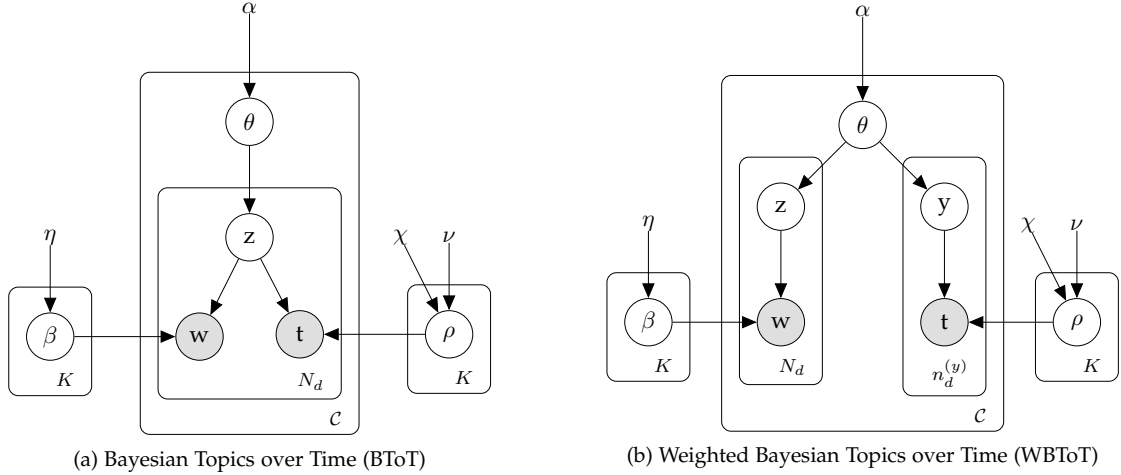
   
\centering
\resizebox{.9\linewidth}{!}{
\begin{minipage}{.5\textwidth}   
\centering
	\tikz{
		\node[const] (alpha) {$\alpha$} ; %
		\node[latent, below=1.25cm of alpha] (theta) {$\theta$} ; %
		\node[latent, below=of theta] (z) {z} ; %
		\node[const, left=2.5cm of z] (eta) {$\eta$} ; %
		\node[latent, below=1cm of eta] (beta) {$\beta$} ; %
		\node[obs, right=1.5cm of beta] (w) {w} ; %

		\node[const, right=2.5cm of z] (nu) {$\nu$} ; %
		\node[latent, below=1cm of nu] (rho) {$\rho$} ; %
		\node[const, left=0.5cm of nu] (chi) {$\chi$} ; %
		\node[obs, left=1.5cm of rho] (t) {t} ; %

		\plate[inner sep=0.25cm, yshift=0.12cm] {plate0} {(beta)} {$K$}; %
		\plate[inner sep=0.25cm, yshift=0.12cm] {plate3} {(rho)} {$K$}; %
		\plate[inner sep=0.25cm, yshift=0.12cm] {plate1} {(z) (w) (t)} {$N_d$}; %
		\plate[inner sep=0.25cm, yshift=0.12cm] {plate2} {(theta) (plate1)} {$\mathcal{C}$}; %
		\edge {alpha} {theta} ; %
		\edge {theta} {z} ; %
		\edge {z,beta} {w} ; %
		\edge {eta} {beta} ; %
		\edge {chi} {rho} ; %
		\edge {nu} {rho} ; %
		\edge {rho} {t} ; %
		\edge {z} {t} ; %
	}

\subcaption{Bayesian Topics over Time (BToT)}   
\label{fig:dag_btot} 
\end{minipage}%
\begin{minipage}{.5\textwidth}
\centering
	\tikz{ %
		\node[const] (alpha) {$\alpha$} ; %
		\node[latent, below=1.25cm of alpha] (theta) {$\theta$} ; %
		\node[latent, below left=of theta] (z) {z} ; %
 	   \node[latent, below right= of theta] (y) {y} ; %

		\node[obs, below=1cm of z] (w) {w} ; %
		\node[latent, left=1.25cm of w] (beta) {$\beta$} ; %
		\node[const, above=1cm of beta] (eta) {$\eta$} ; %
		\node[obs, below=1cm of y] (t) {t} ; %

		\node[latent, right=1.25cm of t] (rho) {$\rho$} ; %
		\node[const, above=1cm of rho] (nu) {$\nu$} ; %
		\node[const, left=0.5cm of nu] (chi) {$\chi$} ; %

		\plate[inner sep=0.25cm, yshift=0.12cm] {plate0} {(beta)} {$K$}; %
		\plate[inner sep=0.25cm, yshift=0.12cm] {plate3} {(rho)} {$K$}; %
		\plate[inner sep=0.25cm, yshift=0.12cm] {plate1} {(z) (w)} {$N_d$}; %
		\plate[inner sep=0.25cm, yshift=0.12cm] {plate4} {(y) (t)} {$n^{(y)}_d$}; %
		\plate[inner sep=0.25cm, yshift=0.12cm] {plate2} {(theta) (plate1) (plate4)} {$\mathcal{C}$}; %
		\edge {alpha} {theta} ; %
		\edge {theta} {z} ; %
		\edge {z,beta} {w} ; %
		\edge {eta} {beta} ; 
		\edge {chi} {rho} ; %
		\edge {nu} {rho} ; %
		\edge {y} {t} ; %
		\edge {theta} {y} ; %
		\edge {rho} {t} ; %
	} 

\subcaption{Weighted Bayesian Topics over Time (WBToT)} 
\label{fig:dag_wbtot} 
\end{minipage}%
}
\caption{Directed acyclic graphs for Bayesian approaches to ToT}
\end{figure}

If one assumes that the Beta distribution parameters in the ToT model
are now drawn from this Beta-prior, then the generative process can
be described as:
\begin{enumerate}
\item Draw $K$ multinomial parameters $\boldsymbol{\beta_{k}}$ from $K$ Dirichlet distributions
with parameters $\boldsymbol{\eta_{k}}$.
\item Draw $K$ Beta distribution parameters $\boldsymbol{\rho}_{k}$
from $K$ Beta-prior distributions with parameters $\nu,\boldsymbol{\chi}$.
\item For each document $d$, draw a set of multinomial parameters $\boldsymbol{\theta_{d}}$ from a symmetric
Dirichlet distribution with parameters $\boldsymbol{\alpha}$. Then, for each word
index $i\in\{1,\ldots, N_{d}\}$ in document $d$:
\begin{enumerate}
\item Draw a topic $z_{di}$ from a multinomial with parameters $\boldsymbol{\theta_{d}}$;
\item Draw a word $w_{di}$ from a multinomial with parameters $\boldsymbol{\beta_{z_{di}}}$;
\item Draw a timestamp $t_{di}$ from a Beta distribution with parameters $\boldsymbol{\rho}_{z_{di}}$.
\end{enumerate}
\end{enumerate}
See Fig. \ref{fig:dag_btot} for the directed acyclic graph representation
of this generative process (it is useful to compare this figure with
the corresponding graph representation of the classical ToT, which
we have reproduced in Fig. \ref{fig:dag_tot}). Unlike ToT, this
generative process is now fully Bayesian, hence the name Bayesian
Topics over Time (BToT).

The log-likelihood for BToT can be written as

\begin{equation}
\mathcal{L}^{\text{BToT}}=\mathcal{L}^{\text{LDA}}+\sum_{d=1}^{D}\sum_{i=1}^{N_{d}}\log\underbrace{p\left(t_{di}|\boldsymbol{\rho}_{z_{di}}\right)}_{\text{Beta}}+\sum_{k=1}^{K}\log\underbrace{p\left(\boldsymbol{\rho}_{k}\left|\nu,\boldsymbol{\chi}\right.\right)}_{\text{Beta-prior}},\label{eq:BToT_logLikehood}
\end{equation}
where $\mathcal{L}^{\text{LDA}}$ is the log-likelihood of the standard
LDA (see Eq. (\ref{eq:LDAlogLikehood})). Notice that the only difference
with respect to the classical ToT log-likelihood in Eq. (\ref{eq:TopicsOverTimeLogLikehood})
is the introduction of the Beta-prior. It is important to note that
the two modalities (words and timestamps) of the BToT log-likelihood
are decoupled, so they can be optimized separately. For reference,
let us flesh out the terms in the log-likelihood that depend on timestamps
and their priors (which we will denote $\mathcal{L^{\text{BToT}}}_{ts}$):
\begin{equation}
\begin{aligned}\mathcal{L^{\text{BToT}}}_{ts} & =\sum_{d=1}^{D}\sum_{i=1}^{N_{d}}\left\{ \left(\rho_{z_{di}}^{1}-1\right)\log t_{di}+\left(\rho_{z_{di}}^{2}-1\right)\log\left[1-t_{di}\right]-\log B\left(\boldsymbol{\rho}_{z_{di}}\right)\right\} \\
 & +\nu\sum_{k=1}^{K}\left\{ \boldsymbol{\rho}_{k}\cdot\boldsymbol{\chi}-\log B\left(\boldsymbol{\rho}_{k}\right)\right\} +K\log f\left(\nu,\boldsymbol{\chi}\right),
\end{aligned}
\label{eq:BToT_ts_logLikehood}
\end{equation}
where we have assigned the topic $z_{di}\in\{1,\ldots, K\}$ and timestamp
$t_{di}\in\left(0,1\right)$ to the word $i$ on document $d$.

\subsubsection{Variational Inference\label{subsubsec:btot-vational-inference}}

It is well-known that the problem of computing the posterior distribution
of the hidden variables given a document is intractable for LDA-based
models. However, a variational approximation can be applied by considering
a family of lower bounds. In the mean-field factorization approach,
we consider a fully factorized distribution $q^{\text{BToT}}$ which
consists of the standard LDA ansatz $q^{\text{LDA}}$ (see Eq. (\ref{eq:LDAVariationalAnsatz}))
times $K$ terms of the form $q_{k}\left(\boldsymbol{\rho}_{k}\right)$:

\begin{equation}
q^{\text{BToT}}=q^{\text{LDA}}\prod_{k}q_{k}\left(\boldsymbol{\rho}_{k}\right).
\end{equation}
Notice that we are not imposing any explicit form for the $q_{k}\left(\boldsymbol{\rho}_{k}\right)$
distributions. The resulting evidence lower bound (ELBO) is 

\begin{equation}
\mathcal{L^{\text{BToT-ELBO}}}=\left\langle \mathcal{L^{\text{BToT}}}-\log q^{\text{BToT}}\right\rangle _{q^{\text{BToT}}}.
\end{equation}
The new terms that appear in the ELBO with respect to the standard
LDA ELBO (see Eq. (\ref{eq:lda_elbo})), $\mathcal{L^{\text{BToT-ELBO}}}_{ts}$,
read
\begin{equation}
\begin{aligned}\mathcal{L^{\text{BToT-ELBO}}}_{ts} & =\left\langle \mathcal{L^{\text{BToT}}}_{ts}-\sum_{k=1}^{K}\log q_{k}\left(\boldsymbol{\rho}_{k}\right)\right\rangle _{q^{\text{BToT}}}\\
 & =\sum_{dw}n_{dw}\sum_{k}\phi_{dwk}\left\{ \left(\left\langle \rho_{k}^{1}\right\rangle _{q_{k}\left(\boldsymbol{\rho}_{k}\right)}-1\right)\log t_{d}+\left(\left\langle \rho_{k}^{2}\right\rangle _{q_{k}\left(\boldsymbol{\rho}_{k}\right)}-1\right)\log\left[1-t_{d}\right]\right.\\
 & \left.-\left\langle \log B\left(\boldsymbol{\rho}_{k}\right)\right\rangle _{q_{k}\left(\boldsymbol{\rho}_{k}\right)}\right\} \\
 & +\nu\sum_{k=1}^{K}\left\{ \left\langle \boldsymbol{\rho}_{k}\right\rangle _{q_{k}\left(\boldsymbol{\rho}_{k}\right)}\cdot\boldsymbol{\chi}-\left\langle \log B\left(\boldsymbol{\rho}_{k}\right)\right\rangle _{q_{k}\left(\boldsymbol{\rho}_{k}\right)}\right\} +K\log f\left(\nu,\boldsymbol{\chi}\right)\\
 & -\sum_{k=1}^{K}\left\langle \log q_{k}\left(\boldsymbol{\rho}_{k}\right)\right\rangle _{q_{k}\left(\boldsymbol{\rho}_{k}\right)},
\end{aligned}
\label{eq:logLikehoodVariational}
\end{equation}
where following the notation in \citet{Hoffman2010} we denote $n_{dw}$
the number of occurrences of word $w$ in document $d$ and we write
the mean field contribution to the topics attributions as $q\left(z_{dw}^{k}=1\left|\boldsymbol{\phi}_{dw}\right.\right)=\phi_{dwk}$.
We have assumed that there is only one timestamp $t_{d}$ per document,
so we have set $t_{di}=t_{d}$.

The extremization of Eq. (\ref{eq:logLikehoodVariational}) with respect
to $q_{k}\left(\boldsymbol{\rho}_{k}\right)$, subject to integrability
constraints $1=\int\int d\boldsymbol{\rho}_{k}q_{k}\left(\boldsymbol{\rho}_{k}\right)$,
is given by
\begin{equation}
\frac{\delta}{\delta q_{k}\left(\boldsymbol{\rho}_{k}\right)}\left\{ \mathcal{L^{\text{BToT-ELBO}}}_{ts}+\xi_{k}\sum_{k}\left(\int\int d\boldsymbol{\rho}_{k}q_{k}\left(\boldsymbol{\rho}_{k}\right)_{k}-1\right)\right\} =0,
\end{equation}
where we have introduced $K$ Lagrange multipliers $\xi_{k}$. This
leads \citep{Bishop2007} to another Beta-prior
distribution with parameters $\mu_{k}$ and $\boldsymbol{\psi}_{k}$
such that
\begin{equation}
\begin{aligned}q_{k}\left(\boldsymbol{\rho}_{k}\right) & =f\left(\mu_{k},\boldsymbol{\psi}_{k}\right)\exp\left[\mu_{k}\left(\boldsymbol{\psi}_{k}\cdot\boldsymbol{\rho}_{k}-\log B\left(\boldsymbol{\rho}_{k}\right)\right)\right]\\
\mu_{k} & =\nu+N_{k}\\
\boldsymbol{\psi}_{k} & =\mu_{k}^{-1}\left(N_{k}\boldsymbol{l}_{k}+\nu\boldsymbol{\chi}\right),
\end{aligned}
\label{eq:variationalBetaPrior}
\end{equation}
where 
\begin{equation}
\begin{aligned}N_{k}= & \sum_{d}N_{dk}\\
N_{dk}= & \sum_{w}n_{dw}\phi_{dwk},
\end{aligned}
\label{eq:WordCounts}
\end{equation}
are topic token average counts, and we have defined the vector of average
log-timestamps as 
\begin{align}
\boldsymbol{l}_{k} & =\left\langle \left(\text{\ensuremath{\log}}t_{d},\text{\ensuremath{\log}}\left(1-t_{d}\right)\right)\right\rangle _{k}:=N_{k}^{-1}\sum_{d}N_{dk}\left(\log t_{d},\log\left(1-t_{d}\right)\right).
\label{eq:log-timestamps}
\end{align}

\begin{table}
\centering{}%
\caption{\label{tbl:Notation_Observed}Observed data and other constants}
\begin{tabular}{cl}
\hline 
Symbol & Description\tabularnewline
\hline 
$w_{di}\in\{1,\ldots, V\}$ & Word corresponding to the $i$th token in document $d$\tabularnewline
$n_{dw}$ & Number of occurrences of word $w$ in document $d$\tabularnewline
$N_{d}$ & Total number of tokens in document $d$\tabularnewline
$D$ & Total number of documents in the corpus\tabularnewline
$V$ & Total number of words in the vocabulary\tabularnewline
$t_{di} = t_{d}\in\left(0,1\right)$ & Timestamp corresponding to the $i$th token in document $d$\tabularnewline
$n_{d}^{\left(y\right)}$ & Number of timestamp observations in document $d$ (in WBToT)\tabularnewline
$K$ & Total number of topics\tabularnewline
\hline 
\end{tabular}
\end{table}

The update to the topic attributions multinomials is changed from
Eq. (\ref{eq:lda_phi_equation}) in standard LDA to
\begin{align}
\phi_{dwk} & \propto\exp\left[\left\langle \log\theta_{dk}\right\rangle _{q^{\text{LDA}}}+\left\langle \log\beta_{kw}\right\rangle _{q^{\text{LDA}}}\right.\nonumber \\
 & \left.+\left(\left\langle \rho_{k}^{1}\right\rangle _{q_{k}\left(\boldsymbol{\rho}_{k}\right)}-1\right)\log t_{d}+\left(\left\langle \rho_{k}^{2}\right\rangle _{q_{k}\left(\boldsymbol{\rho}_{k}\right)}-1\right)\log\left(1-t_{d}\right)-\left\langle \log B\left(\boldsymbol{\rho}_{k}\right)\right\rangle _{q_{k}\left(\boldsymbol{\rho}_{k}\right)}\right].\label{eq:phiUpdate}
\end{align}
Due to the fact that the normalization function $f\left(\mu_{k},\boldsymbol{\psi}_{k}\right)$
in Eq. (\ref{eq:variationalBetaPrior}) cannot be calculated analytically,
the averages $\left\langle \boldsymbol{\rho}_{k}\right\rangle _{q_{k}\left(\boldsymbol{\rho}_{k}\right)}$
and $\left\langle \log B\left(\boldsymbol{\rho}_{k}\right)\right\rangle _{q_{k}\left(\boldsymbol{\rho}_{k}\right)}$
that appear in this expression must be computed numerically. However,
it can be useful to consider an asymptotic approximation of these integrals
in order to build intuition and analyze the differences with respect
to standard ToT. To achieve this, notice that $N_{k}\gg1$\footnote{Unless one proposes a very large number of topics (or a small minibatch
size in the case of online learning).}, so from Eqs. (\ref{eq:variationalBetaPrior}) - (\ref{eq:log-timestamps}) it follows that $\mu_k \sim N_{k}$ and $\boldsymbol{\psi_k} \sim 1$ (we have
assumed a moderate size for the model hyper-parameters, $\boldsymbol{\chi},\nu \sim 1$, and that the order of magnitude notation "$\sim$" may include an arbitrary multiplicative constant).
Therefore, one can use the leading Laplace approximation shown in Appendix
\ref{sec:Asymptotic-expansion-of}, Eqs. (\ref{eq:AsymptoticMomentRho}) -
(\ref{eq:AsymptoticMomentLogB}), which to order $O\left(N_{k}^{-1}\right)$
yields
\begin{equation}
\left\langle \boldsymbol{\rho}_{k}\right\rangle _{q_{k}}\simeq\boldsymbol{\rho}_{k0}\label{eq:rho_asymptotic_approx}
\end{equation}
\begin{equation}
\left\langle \log B\left(\boldsymbol{\rho}_{k}\right)\right\rangle _{q_{k}}\simeq\log B\left(\boldsymbol{\rho}_{k0}\right),\label{eq:eq:log_B_asymptotic_approx}
\end{equation}
 where $\boldsymbol{\rho}_{k0}$ has been defined as 
\begin{equation}
\boldsymbol{\rho}_{k0}=\mathrm{argmax_{\boldsymbol{\rho_{k}}}\left\{ \boldsymbol{\rho_{k}}\cdot\text{\ensuremath{\boldsymbol{\psi_{k}}}}-\log B\left(\boldsymbol{\rho_{k}}\right)\right\} },
\end{equation}
and we have made use of standard big $O$ notation. Note that Eq. (\ref{eq:phiUpdate}) together with Eqs. (\ref{eq:rho_asymptotic_approx})
- (\ref{eq:eq:log_B_asymptotic_approx}) is identical to Eq. (\ref{eq:tot_phi_equation})
in standard ToT. However, from its definition, it is immediate to
see that $\boldsymbol{\rho}_{k0}$ now satisfies the equations
\begin{align}
\frac{1}{1+\nu/N_{k}}\left(\left\langle \text{\ensuremath{\log}}t_{d}\right\rangle _{k}+\frac{\nu}{N_{k}}\chi_{1}\right) & =\Psi\left(\rho_{k0}^{1}\right)-\Psi\left(\rho_{k0}^{1}+\rho_{k0}^{2}\right)\label{eq:rho_k0_eq_1}\\
\frac{1}{1+\nu/N_{k}}\left(\left\langle \text{\ensuremath{\log}}\left(1-t_{d}\right)\right\rangle _{k}+\frac{\nu}{N_{k}}\chi_{2}\right) & =\Psi\left(\rho_{k0}^{2}\right)-\Psi\left(\rho_{k0}^{1}+\rho_{k0}^{2}\right),\label{eq:rho_k0_eq_2}
\end{align}
where $\Psi\left(x\right):=d\log\Gamma\left(x\right)/dx$ is the digamma
function. These equations constitute a regularized version of those
in standard ToT (see Eqs. (\ref{eq:tot_rho1_equation}) - (\ref{eq:tot_rho2_equation})),
which can be recovered by setting $\nu=0$ here. In the unregularized
case, if $1-\left(e^{\left\langle \text{\ensuremath{\log}}t_{d}\right\rangle _{k}}+e^{\left\langle \text{\ensuremath{\log}}\left(1-t_{d}\right)\right\rangle _{k}}\right)\ll1$
(which occurs for a topic $k$ poorly represented in a mini-batch\footnote{Very likely for a large number of topics and/or small mini-batch sizes.}
or with a very peaked timestamp structure\footnote{Common sense indicates that this situation is probably marginal, unless
the mini-batch come with very close timestamps.}), $\boldsymbol{\rho}_{k0}$ could grow by a large factor \citep{Lau1991}.
However, within the present Bayesian approach, the hyper-parameters
$\boldsymbol{\chi}$ and $\nu$ will automatically limit the growth
of $\boldsymbol{\rho}_{k0}$, preventing instabilities. From a practical
point of view, this is the main advantage of BToT with respect to
ToT.

The other two variational parameters that appear in $q^{\text{LDA}}$
(i.e. $\gamma_{dk}$ and $\lambda_{kw}$) are the same as
in standard LDA (see Eqs. (\ref{eq:lda_gamma_equation}) - (\ref{eq:lda_lambda_equation})
in Appendix \ref{sec:Description-of-LDA}), since they are decoupled
from the timestamp part of the log-likelihood. The complete algorithm
for variational inference in BToT is shown in Algorithm \ref{alg:BatchBToT}.

\begin{table*}
\caption{\label{tbl:NotationParameters}Probability distribution parameters and latent variables}

\centering{}%
\resizebox{\textwidth}{!}{
\begin{tabular}{cl}
\hline
Symbol & Description\tabularnewline
\hline 
$\boldsymbol{\theta_d}=\{\theta_{dk}\}_{k=1,\ldots, K}$ & Parameters of the document-topic multinomial distribution\tabularnewline
$\boldsymbol{\alpha}=\left\{ \alpha_{k}\right\} _{k=1,\ldots, K}$ & Parameters of the Dirichlet prior for the topic-document multinomial distribution\tabularnewline
$\boldsymbol{\beta_k}=\{\beta_{kw}\}_{w=1,\ldots, V}$ & Parameters of the topic-word multinomial distribution\tabularnewline
$\boldsymbol{\eta_{k}}=\left\{ \eta_{kw}\right\} _{w=1,\ldots, V}$ & Parameters of the Dirichlet prior for the topic-word multinomial distribution\tabularnewline
$\boldsymbol{\rho}_{k}=\left(\rho_{k}^{1},\rho_{k}^{2}\right)$ & Parameters of the time-topic Beta distribution\tabularnewline
$\nu,\boldsymbol{\chi}=\left(\chi^{1},\chi^{2}\right)$ & Parameters of the Beta-prior prior for the time-topic Beta distribution\tabularnewline
$z_{di}\in\{1,\ldots, K\}$ & Topic assignment to word token $i$ in document $d$\tabularnewline
$y_{d}\in\{1,\ldots, K\}$ & Additional topic assignment to document $d$ in WBToT\tabularnewline
$\boldsymbol{\phi_{dw}}=\left\{ \phi_{dwk}\right\} _{k=1,\ldots, K}$ & Parameters of the variational multinomial distribution for topic assignment of word
$w$ in document $d$\tabularnewline
$\boldsymbol{\gamma_{d}}=\left\{ \gamma_{dk}\right\} _{k=1,\ldots, K}$ & Parameters of the variational Dirichlet prior for the topic-document multinomial distribution\tabularnewline
$\boldsymbol{\lambda_{k}}=\left\{ \lambda_{kw}\right\} _{w=1,\ldots, V}$ & Parameters of the variational Dirichlet prior for the topic-word multinomial distribution\tabularnewline
$\mu_{k},\boldsymbol{\psi}_{k}=\left(\psi_k^{1},\psi_k^{2}\right)$ & Parameters of the variational Beta-prior distribution\tabularnewline
$\boldsymbol{\epsilon_{d}}=\left\{ \epsilon_{dk}\right\} _{k=1,\ldots, K}$ &  Parameters of the variational multinomial distribution for additional topic assignment of document $d$ in WBToT\tabularnewline
\hline 
\end{tabular}}
\end{table*}

\subsubsection{Online optimization}

Online optimization requires the computation of the natural gradient
\citep{Hoffman2010,Hoffman2013} of the one-document contribution to
the\emph{ }ELBO, $l_{d}$. From Eq. (\ref{eq:logLikehoodVariational}), we
see that this contribution is given by 

\begin{align}
l_{d}: & =\sum_{k=1}^{K}N_{dk}\left\{ \left(\left\langle \boldsymbol{\rho}\right\rangle _{q_{k}}-\boldsymbol{1}\right)\cdot\boldsymbol{lt}_{d}-\left\langle \log B\left(\boldsymbol{\rho}\right)\right\rangle _{q_{k}}\right\} \\
 & +\frac{1}{D}\sum_{k=1}^{K}\left\{ \left\langle \boldsymbol{\rho}\right\rangle _{q_{k}}\cdot\left(\nu\boldsymbol{\chi}-\mu_{k}\text{\ensuremath{\boldsymbol{\psi}{}_{k}}}\right)-\left\langle \log B\left(\boldsymbol{\rho}\right)\right\rangle _{q_{k}}\left(\nu-\mu_{k}\right)+\log f\left(\nu,\boldsymbol{\chi}\right)-\log f\left(\mu_{k},\text{\ensuremath{\boldsymbol{\psi}{}_{k}}}\right)\right\} \nonumber ,
\end{align}
where we have defined
\begin{equation}
\boldsymbol{lt}_{d}:=\left(\log t_{d},\log\left[1-t_{d}\right]\right),
\end{equation}
and we have explicitly exhibited the dependence on $\mu_{k}$ and
$\boldsymbol{\psi}_{k}$ by making use of Eq. (\ref{eq:variationalBetaPrior}).
For convenience, we will make use of the ``natural'' variables $\mu'_{k}:=\mu_{k},\boldsymbol{\psi}'_{k}:=\mu_{k}\text{\ensuremath{\boldsymbol{\psi}{}_{k}}}$.
The natural gradients, $\hat{\nabla}$, in these coordinates are
\begin{align}
\hat{\nabla}_{\mu'_{k}}l_{d} & =N_{dk}+\frac{1}{D}\left(\nu-\mu'_{k}\right),\\
\hat{\nabla}_{\boldsymbol{\psi}'_{k}}l_{d} & =N_{dk}\boldsymbol{lt}_{d}+\frac{1}{D}\left(\nu\boldsymbol{\chi}-\boldsymbol{\psi}'_{k}\right).
\end{align}
Given the values of the variables $\mu'_{k}\left(t\right)$ and $\boldsymbol{\psi}'_{k}\left(t\right)$
at iteration $t$, maximization leads to the following online updates
after a mini-bath (MB) of $S$ documents,
\begin{equation}
\mu'_{k}\left(t+1\right)=\left(1-\rho_{t}\right)\mu'_{k}\left(t\right)+\rho_{t}\left(\nu+\frac{D}{S}\sum_{d\in\text{MB}}N_{dk}\right)\label{eq:btot_mu_online_update}
\end{equation}
\begin{equation}
\boldsymbol{\psi}'_{k}\left(t+1\right)=\left(1-\rho_{t}\right)\boldsymbol{\psi}'_{k}\left(t\right)+\rho_{t}\left(\nu\boldsymbol{\chi}+\frac{D}{S}\sum_{d\in\text{MB}}N_{dk}\boldsymbol{lt}_{d}\right)\label{eq:btot_psi_online_update},
\end{equation}
where $\rho_{t}:=\left(t+\tau\right)^{-\kappa}$ is the standard ``mixing''
parameter \citep{Hoffman2010,Hoffman2013}. Here, $\kappa \in \left(0.5,1\right]$ regulates how previous values of the parameters are forgotten, and $\tau \geq 0$ slows down the first iterations.

The complete algorithm is shown in Algorithm \ref{alg:OnlineBToT}.

\begin{algorithm}
\caption{\label{alg:BatchBToT} Batch variational BToT}

\begin{algorithmic}
\STATE Initialize variational parameters $\gamma_{dk},\lambda_{kw},\mu_k,\boldsymbol{\psi}_k$.
\REPEAT
\STATE \emph{E-step:}
\FOR{$d=1$ to $D$}
\REPEAT
\STATE Set $\phi_{dwk}$ from Eq. (\ref{eq:phiUpdate}).
\STATE Set $\gamma_{dk}$ from Eq. (\ref{eq:lda_gamma_equation}).
\UNTIL $\gamma_{dk}$ converged.
\ENDFOR
  
\STATE \emph{M-step:}
\STATE Set $\lambda_{kw}$ from Eq. (\ref{eq:lda_lambda_equation}).
\STATE Set $\mu_k,\boldsymbol{\psi}_k$ from Eq. (\ref{eq:variationalBetaPrior}).

\UNTIL $\mathcal{L}$ converged. 
\end{algorithmic}
\end{algorithm}

\begin{algorithm}
\caption{\label{alg:OnlineBToT} Online variational BToT}
\begin{algorithmic}

\STATE Set $t=0$.
\STATE Initialize variational parameters $\lambda_{kw}(t),\mu_k(t),\boldsymbol{\psi}_k(t)$.
\REPEAT
\STATE Set $t \gets t+1$.
\STATE Set $\rho_{t}:=\left(t+\tau\right)^{-\kappa}$.
\STATE \emph{E-step:}

\FOR{$d=1$ to $S$}
\STATE Initialize variational parameters $\gamma_{dk}$.
\REPEAT
\STATE Set $\phi_{dwk}$ from Eq. (\ref{eq:phiUpdate}).
\STATE Set $\gamma_{dk}$ from Eq. (\ref{eq:lda_gamma_equation}).
\UNTIL {$\gamma_{dk}$ converged}.
\ENDFOR
\STATE \emph{M-step:}
\STATE Set $\lambda_{kw}(t+1)$ from Eq. (\ref{eq:lda_lambda_equation_online}).
\STATE Set $\mu_k(t+1),\boldsymbol{\psi}_k(t+1)$ from Eq. (\ref{eq:btot_mu_online_update}) and Eq. (\ref{eq:btot_psi_online_update}).
\UNTIL $\mathcal{L}$ converged.
\end{algorithmic}
\end{algorithm}

\subsection{Weighted Bayesian ToT (WBToT)\label{subsec:WBToT}}

\subsubsection{Description of the model\label{subsubsec:wbtot-description}}

Notice that, although a single timestamp, $t_{d}$, is observed for
each document $d$, the BToT model sees it $N_{d}$ times during
training, since we generate a timestamp for each word of the document.
As we will empirically see in Section \ref{sec:experimental-results},
this places excessive importance on timestamps while relegating
semantic coherence of topics to a secondary place. To balance the
relative weight between words and timestamps and rescale both modalities
of the log-likelihood, a balancing hyperparameter is typically introduced
\citep{Wang2006,Fang2017}.

However, as exposed in Appendix \ref{sec:Balancing-hyper-parameter},
it is difficult to follow this approach from a Bayesian perspective.
To solve this problem, we slightly modify BToT to mimic the introduction
of this balancing hyperparameter while keeping the Bayesian approach.
We will call this model Weighted BToT (WBToT). In WBToT, an additional
latent topic assignment variable, $y_{d}\in\{1,\ldots,K\}$, is introduced
to account for the generation of timestamp observations from topics,
instead of using the same $z_{di}$ we employed for word generation.
These new topic assignments are drawn from the same multinomial distributions
as $z$, i.e. $p\left(y_{d}|\theta_{d}\right)=\theta_{dy_{d}}$. The
observation of the document's timestamp through $y$ is repeated $n_{d}^{(y)}$
times to mimic the effect of the balancing hyperparameter. This way,
we can effectively control the relative weight of the two modalities
of the log-likelihood.

The generative process for the WBToT model is as follows:
\begin{enumerate}
\item Draw $K$ multinomial parameters $\boldsymbol{\beta_{k}}$ from $K$ Dirichlet distributions
with parameters $\boldsymbol{\eta_{k}}$.
\item Draw $K$ Beta distribution parameters $\boldsymbol{\rho}_{k}$
from $K$ Beta-prior distributions with parameters $\nu,\boldsymbol{\chi}$.
\item For each document $d$, draw a set of multinomial parameters $\boldsymbol{\theta_{d}}$ from a symmetric
Dirichlet distribution with parameters $\boldsymbol{\alpha}$.
\begin{enumerate}
\item Then, for each word index $i\in\{1,\ldots, N_{d}\}$ in document $d$:
\begin{enumerate}
\item Draw a topic $z_{di}$ from a multinomial with parameters $\boldsymbol{\theta_{d}}$;
\item Draw a word $w_{di}$ from a multinomial with parameters $\boldsymbol{\beta_{z_{di}}}$;
\end{enumerate}
\item Draw $n_{d}^{\left(y\right)}$ topics $y_{d}^{i}$, with $i\in\{1,\ldots, n_{d}^{\left(y\right)}\}$, from a multinomial with parameters $\boldsymbol{\theta_{d}}$. For each of them:
\begin{enumerate}
\item Draw a timestamp $t_{d}^{i}$ from a Beta distribution with parameters
$\boldsymbol{\rho}_{y_{d}^{i}}$.
\end{enumerate}
\end{enumerate}
\end{enumerate}
See Fig. \ref{fig:dag_wbtot} for the directed acyclic graph representation
of this generative process. The introduction of these repeated timestamp
observations allows balancing the relative influence of words and
timestamps along the inference process. For example, $n_{d}^{\left(y\right)}=1$
corresponds strictly to the Bayesian version of the alternative ToT
model depicted in Fig. \ref{fig:dag_tot_alternate} and $n_{d}^{\left(y\right)}=\delta N_{d}$
approximately corresponds to a Bayesian version of the ToT model with
balance hyperparameter $\delta$ introduced in \citet{Fang2017}.

The resulting log-likelihood for WBToT is

\begin{equation}
\mathcal{L}^{\text{WBToT}}=\mathcal{L}^{\text{LDA}}+\sum_{d=1}^{D}\sum_{i=1}^{n_{d}^{\left(y\right)}}\left\{ \log\underbrace{p\left(t_{d}^{i}|\boldsymbol{\rho}_{y_{d}^{i}}\right)}_{\textrm{Beta}}+\log\underbrace{p\left(y_{d}^{i}|\boldsymbol{\theta_{d}}\right)}_{\textrm{Multinomial}}\right\} +\sum_{k=1}^{K}\log\underbrace{p\left(\boldsymbol{\rho}_{k}|\nu,\boldsymbol{\chi}\right)}_{\textrm{Beta-prior}}.\label{eq:WBToT_logLikehood}
\end{equation}
For reference, let us flesh out the terms in the log-likelihood that
depend on timestamps and their priors (which we will denote $\mathcal{L^{\text{WBToT}}}_{ts}$):

\begin{equation}
\begin{aligned}\mathcal{L}_{ts}^{\text{WBToT}}= & \sum_{d=1}^{D}\sum_{i=1}^{n_{d}^{\left(y\right)}}\left\{ \left(\rho_{y_{d}^{i}}^{1}-1\right)\log t_{d}^{i}+\left(\rho_{y_{d}^{i}}^{2}-1\right)\log\left[1-t_{d}^{i}\right]-\log B\left(\boldsymbol{\rho}_{y_{d}^{i}}\right)+\log\theta_{dy_{d}^{i}}\right\} \\
 & +\nu\sum_{k=1}^{K}\left\{ \boldsymbol{\rho}_{k}\cdot\boldsymbol{\chi}-\log B\left(\boldsymbol{\rho}_{k}\right)\right\} +K\log f\left(\nu,\boldsymbol{\chi}\right).
\end{aligned}
\label{eq:WBToT_ts_logLikehood}
\end{equation}

\subsubsection{Variational Inference\label{subsubsec:wbtot-vational-inference}}

The variational ansatz is of the form
\begin{equation}
\begin{aligned}q^{\text{WBToT}}= & q^{\text{LDA}}\prod_{d=1}^{D}\left(\prod_{i=1}^{n_{d}^{\left(y\right)}}\underbrace{q\left(y_{d}^{i}|\boldsymbol{\epsilon_{d}^{i}}\right)}_{\text{Multinomial}}\right)\prod_{k=1}^{K}\underbrace{q_{k}\left(\boldsymbol{\rho}_{k}|\mu_{k},\boldsymbol{\psi}_{k}\right)}_{\text{Beta-prior}},\end{aligned}
\label{eq:ToT2Ansatz}
\end{equation}
where $q^{\text{LDA}}$ is given by Eq. (\ref{eq:LDAVariationalAnsatz}), and we have introduced a new set of
multinomial variational parameters $\epsilon^i_{dk}$.
The resulting ELBO is 
\begin{equation}
\mathcal{L^{\text{WBToT-ELBO}}}=\left\langle \mathcal{L^{\text{WBToT}}}-\log q^{\text{WBToT}}\right\rangle _{q^{\text{WBToT}}}.
\end{equation}
The new terms that appear in the ELBO with respect to the standard
LDA ELBO (see Eq. (\ref{eq:lda_elbo})), $\mathcal{L^{\text{WBToT-ELBO}}}_{ts}$,
read
\begin{equation}
\begin{aligned}\mathcal{L^{\text{WBToT-ELBO}}}_{ts} & =\left\langle \mathcal{L^{\text{WBToT}}}_{ts}-\sum_{d=1}^{D}\sum_{i=1}^{n_{d}^{\left(y\right)}}\log q\left(y_{d}^{i}\right)-\sum_{k=1}^{K}\log q_{k}\left(\boldsymbol{\rho}_{k}\right)\right\rangle _{q^{\text{WBToT}}}\\
 & =\sum_{d=1}^{D}\sum_{i=1}^{n_{d}^{\left(y\right)}}\sum_{k}^{K}\epsilon_{dk}^{i}\left\{ \left(\left\langle \boldsymbol{\rho}_{k}\right\rangle _{q_{k}\left(\boldsymbol{\rho}_{k}\right)}-\boldsymbol{1}\right)\cdot\boldsymbol{lt}_{d}-\left\langle \log B\left(\boldsymbol{\rho}_{k}\right)\right\rangle _{q_{k}\left(\boldsymbol{\rho}_{k}\right)}\right.\\
 & \left.+\left\langle \log\theta_{dk}\right\rangle _{q^{\text{LDA}}}-\log\epsilon_{dk}^{i}\right\} \\
 & +\sum_{k=1}^{K}\left\{ \left\langle \boldsymbol{\rho}_{k}\right\rangle _{q_{k}\left(\boldsymbol{\rho}_{k}\right)}\cdot\left(\nu\boldsymbol{\chi}-\mu_{k}\boldsymbol{\psi}_{k}\right)-\left\langle \log B\left(\boldsymbol{\rho}_{k}\right)\right\rangle _{q_{k}\left(\boldsymbol{\rho}_{k}\right)}\left(\nu-\mu_{k}\right)\right.\\
 & \left.+\log f\left(\nu,\boldsymbol{\chi}\right)-\log f\left(\mu_{k},\boldsymbol{\psi}_{k}\right)\right\} ,
\end{aligned}
\label{eq:logLikehoodVariational-1}
\end{equation}
where the average $\left\langle \log\theta_{dk}\right\rangle _{q^{\text{LDA}}}$
is given by Eq. (\ref{eq:theta_q_mean}), as in standard LDA. Notice that,
due to the introduction of $\gamma_{dk}$ through this term in the
ELBO, the update equation for this variational parameter will be modified
(this is a novelty with respect to BTOT, where $\gamma_{dk}$ is the
same as in LDA). On the other hand, $\phi_{dwk}$ has been replaced
by $\epsilon_{dk}$ everywhere in this part of the ELBO.

Extremization of $\mathcal{L^{\text{WBToT-ELBO}}}$ (subject to the
new constraint $\sum_{k}\epsilon_{dk}^{i}=1$) leads to the following
document specific variational parameters updates 
\begin{equation}
\epsilon_{dk}:=\epsilon_{dk}^{i}\propto\exp\left[\left\langle \log\theta_{dk}\right\rangle _{q^{\text{LDA}}}+\left(\left\langle \boldsymbol{\rho}_{k}\right\rangle _{q_{k}\left(\boldsymbol{\rho}_{k}\right)}-\boldsymbol{1}\right)\cdot\boldsymbol{lt}_{d}-\left\langle \log B\left(\boldsymbol{\rho}_{k}\right)\right\rangle _{q_{k}\left(\boldsymbol{\rho}_{k}\right)}\right]\label{eq:wbtot_epsilon_update}
\end{equation}
\begin{equation}
\gamma_{dk}=\alpha_{k}+\underbrace{\sum_{w}n_{dw}\phi_{dwk}}_{N_{dk}}+n_{d}^{\left(y\right)}\epsilon_{dk},\label{eq:wbtot_gamma_update}
\end{equation}
where we have explicitly noted that $\epsilon_{dk}^{i}$ does not
depend on its index $i$ because there is a single timestamp
observation per document. From these formulae, it follows that the choice
$n_{d}^{\left(y\right)}=1$ behaves as a single word added to the
document-topic pseudo-counts $\gamma_{dk}$, and therefore the timestamp
influence will be small. Since $\phi_{dwk}$ is not present in $\mathcal{L^{\text{WBToT-ELBO}}}_{ts}$,
its update equation will be the same as in LDA, Eq. (\ref{eq:lda_phi_equation}).

Corpus-wide variational parameters obey the following self-consistency
relations:
\begin{equation}
\mu_{k}=\nu+\sum_{d}n_{d}^{\left(y\right)}\epsilon_{dk}\label{eq:wbtot_batch_mu_update}
\end{equation}
\begin{equation}
\boldsymbol{\psi}{}_{k}=\mu_{k}^{-1}\left(\nu\boldsymbol{\chi}+\sum_{d}n_{d}^{\left(y\right)}\epsilon_{dk}\boldsymbol{lt}_{d}\right),\label{eq:wbtot_batch_psi_update}
\end{equation}
and the $\lambda_{kw}$ variational parameter is exactly the same
as in standard LDA (see Eq. (\ref{eq:lda_lambda_equation}) in Appendix \ref{sec:Description-of-LDA}),
since it is decoupled from the timestamp part of the log-likelihood.

It is interesting to note that the variational parameters satisfy
$e^{\psi^{1}}+e^{\psi^{2}}<\left(e^{\chi^{1}}+e^{\chi^{2}}\right)^{\nu/\mu}$(the
$k$-dependence is implicit). This can be proved by using the generalized
Hölder inequality. If we assume $n_{d}^{\left(y\right)}\gg1$, Eqs. (\ref{eq:wbtot_batch_mu_update})
- (\ref{eq:wbtot_batch_psi_update}) imply that the inequality will
be close to saturation. As a consequence, for non very strong Beta-prior
parameters (namely $\nu\lesssim1$), the Beta-prior variational parameters
will be close to the boundary of their validity regions, $1-\left(e^{\psi^{1}}+e^{\psi^{2}}\right)\ll1$.

The updates to the Beta-prior hyper-parameters are the same as in BToT.
The complete algorithm is shown in Algorithm \ref{alg:BatchWBToT}.

\subsubsection{Online optimization}

In the case of online updates, it is immediate to see that one only
has to substitute $N_{dk}\rightarrow n_{d}^{\left(y\right)}\epsilon_{dk}$
in Eqs. (\ref{eq:btot_mu_online_update}) - (\ref{eq:btot_psi_online_update}),
that is:
\begin{equation}
\mu'_{k}\left(t+1\right)=\left(1-\rho_{t}\right)\mu'_{k}\left(t\right)+\rho_{t}\left(\nu+\frac{D}{S}\sum_{d\in MB}n_{d}^{\left(y\right)}\epsilon_{dk}\right),\label{eq:wbtot_online_mu_update}
\end{equation}
\begin{equation}
\boldsymbol{\psi}'_{k}\left(t+1\right)=\left(1-\rho_{t}\right)\boldsymbol{\psi}'_{k}\left(t\right)+\rho_{t}\left(\nu\boldsymbol{\chi}+\frac{D}{S}\sum_{d\in MB}n_{d}^{\left(y\right)}\epsilon_{dk}\boldsymbol{lt}_{d}\right).\label{eq:wbtot_online_psi_update}
\end{equation}
The complete algorithm for online optimization is shown in Algorithm \ref{alg:OnlineWBToT}.
\begin{algorithm}
\caption{\label{alg:BatchWBToT} Batch variational WBToT}
\begin{algorithmic}

\STATE Initialize variational parameters $\gamma_{dk},\lambda_{kw},\mu_k,\boldsymbol{\psi}_k,\epsilon_{dk}$.
\REPEAT
\STATE \emph{E-step:}
\FOR{$d=1$ to $D$}
\REPEAT
\STATE Set $\phi_{dwk}$ from Eq. (\ref{eq:lda_phi_equation}).
\STATE Set $\epsilon_{dk}, \gamma_{dk}$ from Eq. (\ref{eq:wbtot_epsilon_update}) and Eq. (\ref{eq:wbtot_gamma_update}).
\UNTIL $\gamma_{dk}$ converged.
\ENDFOR
\STATE \emph{M-step:}
\STATE Set $\lambda_{kw}$ from Eq. (\ref{eq:lda_lambda_equation}).
\STATE Set $\mu_k,\boldsymbol{\psi}_k$ from Eq. (\ref{eq:wbtot_batch_mu_update}) and Eq. (\ref{eq:wbtot_batch_psi_update}). 
 \UNTIL $\mathcal{L}$ converged.
\end{algorithmic}
\end{algorithm}

\begin{algorithm}
\caption{\label{alg:OnlineWBToT} Online variational WBToT}
\begin{algorithmic}
\STATE Set $t=0$.
\STATE Initialize variational parameters $\lambda_{kw}(t),\mu_k(t),\boldsymbol{\psi}_k(t)$.
\REPEAT
\STATE Set $t \gets t+1$.
\STATE Set $\rho_{t}:=\left(t+\tau\right)^{-\kappa}$.
\STATE \emph{E-step:}
\FOR{$d=1$ to $D$}
\STATE Initialize variational parameters $\gamma_{dk},\epsilon_{dk}$.
\REPEAT
\STATE Set $\phi_{dwk}$ from Eq. (\ref{eq:lda_phi_equation}).
\STATE Set $\epsilon_{dk}, \gamma_{dk}$ from Eq. (\ref{eq:wbtot_epsilon_update}) and Eq. (\ref{eq:wbtot_gamma_update}).
\UNTIL $\gamma_{dk}$ converged.
\ENDFOR
\STATE \emph{M-step:}
\STATE Set $\lambda_{kw}(t+1)$ from Eq. (\ref{eq:lda_lambda_equation_online}).
\STATE Set $\mu_k(t+1),\boldsymbol{\psi}_k(t+1)$ from Eq. (\ref{eq:wbtot_online_mu_update}) and Eq. (\ref{eq:wbtot_online_psi_update}).
\UNTIL $\mathcal{L}$ converged.
\end{algorithmic}
\end{algorithm}

\subsection{Time complexity analysis\label{subsec:time_complexity}}

In this section, we provide the time complexity analysis of the two models introduced in our paper. For reference, we also include LDA and standard ToT in our analysis.

Let $N_{ud}$ be the number of unique tokens in document $d$, $N_g$ the number of iterations in the E-step until $\gamma_{dk}$ convergence, $N_r$ the number of iterations to find roots in ToT E-step, and $T$ the number of unique document timestamps.

It must be noted that $N_g$ and $N_r$ depend very weakly on the corpus and the number of topics. Strictly, they should not be included in the complexity computations. However, they usually amount to $10^1-10^3$, which can be quite significant, typically on the order or larger than the number of topics.

The complexities of variational inferences in the different models are:

\begin{enumerate}
\item LDA: each E-step requires setting $\phi_{dwk}$, which needs $O(N_{ud} \times K)$ operations. This has to be iterated until convergence (of $\gamma_{dk}$), which requires $N_g$ steps. The resulting complexity is $O\left(N_g \times D \times N_{ud} \times K\right)$. The M-step requires $O(D \times N_{ud} \times K + V \times K) = O(D \times N_{ud} \times K)$ operations, which is smaller than the E-step complexity.

\item ToT: on top of LDA complexity, there is another $O(N_r \times D \times N_{ud} \times K)$ contribution to complexity in the E- step. The contribution to the M-step can be neglected. The resulting complexity is $O((N_g + N_r) \times D \times N_{ud} \times K)$.

\item BToT and WBToT: there is a $O(T \times K)$ added complexity to LDA's complexity in E-Step. The resulting complexity is $O(N_g \times D \times N_{ud} \times K) + O(T \times K)$.
\end{enumerate}

\section{Datasets and experimental set-up\label{sec:datasets}}

We will employ two datasets to validate our models. The datasets are
very different in size and nature, and each of them will be used for
different purposes. The first dataset consists of the annual State
of the Union Addresses (SOTU) by US Presidents from 1790 to 2022,
and will be used to compare different topic models between them. The 
second dataset is a massive corpus of tweets about the COVID-19 pandemic
from March 2020 to June 2020, and will be employed to illustrate the
use of the online algorithm in WBToT for a case that may otherwise
be intractable in a batch setting due to memory constraints.

\subsection{State of the Union Addresses\label{subsec:sotu_dataset}}

\subsubsection{Dataset}

As in the original work that introduced the ToT model \citep{Wang2006},
we will use the State of the Union (SOTU) Addresses as our main dataset.
The SOTU is an annual speech given by the president of the United
States describing the economic, political, and social state of the
country. We will use an updated dataset made of the 236 addresses
given in the range 1790-2022 (from George Washington to Joe Biden).
As in \citet{Wang2006}, we split each address into 3-paragraph documents,
which leaves us with a corpus of 7224 documents written in English.
For each of these documents, the publication timestamp is the year
in which the address was given.

We perform a standard text preprocessing on this corpus using the
stanza package \citep{Qi2020}: we tokenize the documents, remove stopwords
and numbers, lemmatize the tokens, and filter out words that appear
in less than 15 documents or in more than 50\% of the documents. This
leaves us with a vocabulary of 4086 words. Finally, we create a bag-of-words
representation of the corpus. The dataset statistics are summarized in Table \ref{tbl:Datasets_Statistics}.

\subsubsection{Experimental set-up}

This is a relatively small dataset that we can employ to compare different
topic models between them since training time is restrained. We will train LDA, BERTopic, BToT, and two WBToT
models on this dataset (the two WBToT models will differ on the $n_{d}^{\left(y\right)}$
parameter, with one of them having $n_{d}^{\left(y\right)}=0.1N_{d}$
and the other $n_{d}^{\left(y\right)}=\sqrt{N_{d}}$). In all cases, we will look
for 50 topics (for BERTopic, this implies reducing the original number of topics by iterative merging).

In probabilistic topic models (i.e LDA, BToT, WBToT), we will optimize the standard hyper-parameters $\boldsymbol{\alpha}$
and $\boldsymbol{\eta_k}$ (initialized to $1/K$), but not the novel ones here introduced,
$\nu$ and $\boldsymbol{\chi}$ (which will be initialized to $1/K$ and $\log0.45\left(1,2\right)$
respectively). It is well known that topic models reach different
optimal values depending on the parameter initialization. To alleviate
this problem, for each model we will run 10 batch simulations with
different parameter initializations, setting as our stopping criterion
that the change in perplexity, $\mathcal{P}^{t}\coloneqq e^{-\mathcal{L}^{t}/N_{tok}}$,
is $\mathcal{P}^{t-1}-\mathcal{P}^{t}<10^{-5}$ (where $N_{tok}$
is the total number of tokens in the corpus and $\mathcal{L}^{t}$
is the ELBO after $t$ iterations of the EM algorithm). We will also establish a maximum limit of 400 EM iterations, so the algorithm is stopped if the convergence criterion has not been satisfied by then. For each topic
model, we will choose the simulation with the largest ELBO out of
the 10 trials as our final result. In the WBToT models, the two options
$n_{d}^{\left(y\right)}=0.1N_{d}$ and $n_{d}^{\left(y\right)}=\sqrt{N_{d}}$
are chosen so that on average the timestamps contribute with equal
weight (here $\left\langle N_{d}\right\rangle =89.7$ and $\left\langle \sqrt{N_{d}}\right\rangle =8.95$),
but $n_{d}^{\left(y\right)}=\sqrt{N_{d}}$ weights timestamps more
in documents with $N_{d}<100$. For BERTopic, we will use the default parameters. The results for this experiment are presented in Section \ref{subsec:wbtot_vs_lda} - \ref{subsec:wbtot_vs_btot}.

The absence of $\nu$ and $\boldsymbol{\chi}$ optimization requires some clarification. We first note that, as explained in Section \ref{sec:Introduction}, the main reasons to introduce the Beta-prior were to obtain a stable online variational inference algorithm and to provide a principled approach to regularizing large gradients. The other usual reason for introducing a prior, overfitting, is not relevant in our case because the Beta distribution won't overfit, as stated in \citet{Wang2006}. When we tried to optimize $\nu$ and $\boldsymbol{\chi}$, we noted that both parameters migrated to the boundaries of their validity region. This behavior removes the regularizing effect that is required to handle large gradients. We guess that, because of the lack of overfitting, the system just tries to make the regularizing term as small as possible. A more detailed study of the optimization of $\nu$ and $\boldsymbol{\chi}$ is left for future research.

\subsection{Large-Scale COVID-19 Twitter Chatter\label{subsec:covid_dataset}}

\subsubsection{Dataset}

The Large-Scale COVID-19 Twitter
Chatter dataset introduced by \citep{Banda2021}. This dataset consists
of over 1.12 billion tweets related to COVID-19 chatter in different
languages. Data was collected from the Twitter Stream API filtering
by keywords such as ``coronavirus'', ``2019nCoV'', ``COVD19'',
``CoronavirusPandemic'', ``COVID-19'', ``CoronaOutbreak'', and
other sensitive terms, plus additional collections of tweets provided by
other researchers (we refer the reader to \citet{Banda2021} for a
detailed description of the dataset). Misinformation in social media during the COVID-19 pandemic has attracted considerable attention from the ML community \citep{Burel2021}, so this is a compelling dataset to test our model on. In particular, we have restricted
ourselves to tweets written in English that are not retweets from
12 March 2020 to 12 June 2020, comprising a three months period. Furthermore,
we have only sampled 10 million random tweets from this subset for efficiency purposes.

We have performed a standard text preprocessing on this corpus using
the NLTK \citep{nltk_book} TweetTokenizer class. As in the previous
dataset, we have lemmatized the documents and removed stopwords and
numbers. Additionally, we have removed Twitter usernames,
hashtags, and URLs. Words that appear in less than
5000 tweets or more than 50\% of the tweets are removed. This leaves
us with a vocabulary of 2611 words. For each tweet, the publication
timestamp is the day on which it was published. The dataset statistics are summarized in Table \ref{tbl:Datasets_Statistics}.

\subsubsection{Experimental set-up}

This is a large dataset that we will employ to illustrate the ability
of WBToT to deal with massive corpora in an online fashion. Its large size makes it unsuitable for model comparisons, but it provides a good scenario to study the evolution of a model as successive batches of data arrive. The dataset
will be shuffled and 10\% of the corpus will be set apart for testing
purposes: the goal is to show that, for each optimization step taken after receiving a mini-batch of data, the held-out perplexity on this test set decreases. We will train a WBToT model with $n_{d}^{\left(y\right)}=\sqrt{N_{d}}$
and a batch size of 1 million documents (which implies that 9 optimizations
will be performed for each iteration of the EM algorithm). As in the SOTU experiment, $\nu$ and $\boldsymbol{\chi}$ are not optimized. The number
of topics will be 10 and only one EM iteration will be executed,
since we are interested in studying the effect of successive batches of data, not the convergence of the EM algorithm itself.

Results for this experiment are presented in Section \ref{subsec:wbtot_online_version}.

\begin{table}
\caption{\label{tbl:Datasets_Statistics}Dataset Statistics}

\centering{}%
\begin{tabular}{lcc}
\hline 
 & SOTU addresses & Twitter\tabularnewline
\hline 
Dataset size & $7224$ documents & $10^7$ documents\tabularnewline
Vocabulary size & $4086$ words & $2611$ words\tabularnewline
Timespan & $332$ years & $3$ months\tabularnewline
Granularity & $1$ year & $1$ day\tabularnewline
$\left\langle N_{d}\right\rangle$ & $89.7$ words & $6.89$ words \tabularnewline
\hline 
\end{tabular}
\end{table}

\section{Results and Implications\label{sec:experimental-results}}

In this section, we present the results of the experiments described
above.  Recall that the main advantage of introducing a time modality into the log-likelihood is the ability to capture events: we will study this property in Section \ref{subsec:wbtot_vs_lda}. However, this time modality could degrade the model's word modeling performance: for this reason, we will study coherence in Section \ref{subsec:wbtot_vs_btot}. Finally, we will illustrate the stability of WBToT in Section \ref{subsec:wbtot_online_version}.

It must be noted BToT generates very low-quality topics, as already discussed in Section \ref{subsec:WBToT}. We observed that, in many cases, the model generates completely uniform word distributions that are impossible to interpret. For this reason, we have not included BToT results in Section \ref{subsec:wbtot_vs_lda}, and further motivation for this will be analyzed in the discussion about coherence in Section \ref{subsec:wbtot_vs_btot}.

\subsection{Event modeling\label{subsec:wbtot_vs_lda}}

Due to the introduction of a time modality in the log-likelihood,
events are better captured in WBToT than in time-unaware topic models (where no information
about the timestamp is provided during training). This implies two
things: first, that topic presence over time is more concentrated
in WBToT, since topics describe particular events rather
than occurred in a specific period of time; and second, that topics
are semantically more distinct between them in WBToT,
since a more singular vocabulary is necessary to describe specific
events. Although this feature was already present in standard ToT \citep{Wang2006}, it is important to note that we can reproduce these results using a fully Bayesian approach. To demonstrate this, we will use the topics extracted in the
SOTU dataset (see Section \ref{subsec:sotu_dataset}) for LDA, BERTopic and
the two versions of WBToT.

\subsubsection{Topic concentration in time}

Topic presence over time can be defined as the number of tokens assigned
to each topic per unit of time (that is, $N_{kt}\coloneqq\sum_{d}^{D}\sum_{w}^{V}\delta_{tt_{d}}\theta_{dk}n_{dw}$,
where $t_{d}$ is the publication timestamp of document $d$ and $\delta_{tt_{d}}$
is a Kronecker delta of $t$ and $t_{d}$). In order to prove that
topics are more concentrated in time in WBToT, we have
to introduce a measure of the dispersion of $N_{tk}$ for a fixed
$k$. Due to the presence of outliers, which
distorts the meaning of typical measures of variability such as the
standard deviation, we will make use of robust metrics of statistical
dispersion: the median absolute deviation (MAD) and the interquartile
range (IQR). The MAD is defined as the median of the absolute deviations from the data's median $\tilde{N_{k}}=\text{median}\left(N_{tk}\right)$,
\begin{equation}
\text{MAD} = \text{median}\left(\left|N_{tk}-\tilde{N_{k}}\right|\right).\label{eq:mad_definition}
\end{equation}
The IQR is defined as the difference between the $75$th percentile, $Q_3$, and the $25$th percentile, $Q_1$,
\begin{equation}
\text{IQR} = Q_3 - Q_1.\label{eq:iqr_definition}
\end{equation}
For each model that we want to compare,
we will compute these metrics for each of the $K$ topics, and then
we will take its mean value. Results are shown in Table \ref{tbl:wbtot_vs_lda_statistics}.

\begin{table}

\begin{centering}
\caption{Robust dispersion statistics for each model\label{tbl:wbtot_vs_lda_statistics}}
\begin{tabular}{ccc}
\hline 
Model & Median Absolute Deviation (MAD) & Interquartile Range (IQR)\tabularnewline
\hline 
$\text{WBToT}(n_{d}^{y}=0.1N_{d})$ & $9.36$ years & $23.37$ years\tabularnewline
$\text{WBToT}(n_{d}^{y}=\sqrt{N_{d}})$ & $11.76$ years & $27.25$ years\tabularnewline
$\text{LDA}$ & $19.27$ years & $41.87$ years\tabularnewline
$\text{BERTopic}$ & $14.18$ years & $31.58$ years\tabularnewline
\hline 
\end{tabular}
\par\end{centering}
\end{table}

It is clear that topic dispersion is greater in LDA than in both WBToT
models. In particular, we observe that the MAD for the WBToT models
is approximately a decade, while for LDA it goes up to almost two
decades. The IQR comprises between two and three decades for the WBToT
models, while it goes beyond four decades for the LDA. These results
imply that topics in LDA are almost half as concentrated in time as
topics in WBToT models. BERTopic seems to be more concentrated in time than LDA, but it
is still far from WBToT. Differences between the two WBToT models are
not very significant, although it seems that the model with $n_{d}^{y}=0.1N_{d}$
is slightly more concentrated in time than the model with $n_{d}^{y}=\sqrt{N_{d}}$.

\subsubsection{Topic semantic distance}

To prove this point, first, we have to introduce a notion of semantic distance between topics. We will employ the symmetric Kullback-Leibler divergence,
defined as 
\begin{equation}
d_{KL}^{\text{sym}}\left(P,Q\right)=D_{KL}\left(P||Q\right)+D_{KL}\left(Q||P\right),\label{eq:sym_kl_div}
\end{equation}
where $D_{KL}\left(P||Q\right)$ is the Kullback-Leibler divergence
of distribution $P$ from distribution $Q$. We will compute the symmetric
Kullback-Leibler divergence between the word distributions $\beta_{kw}$ for all the possible pairs of topics
for each of the three models (from which we obtain $1225$ distances
in each case) and then we will take its mean value. 

Notice that this metric is not well-defined for BERTopic, since the Kullback-Leibler divergence requires that, for all $x$, if $Q(x)=0$, then $P(x)=0$. Probabilistic topic models exhibit a smooth topic distribution thanks to the introduction of priors, which prevents them from suffering from this problem. However, BERTopic makes use of TF-IDF to obtain word distributions, and this implies that the distributions may contain zeros. Therefore, we will exclude BERTopic from this analysis. 

Results are presented
in Table \ref{tbl:wbtot_vs_lda_distances}. Both WBToT models show a higher intra-model semantic distance between pairs of topics than LDA. This result is due to the fact that WBToT separates topics corresponding to different periods of time, and documents published during the same time span tend to share a similar vocabulary.

\begin{table}
\caption{Mean semantic distance for each model\label{tbl:wbtot_vs_lda_distances}}

\centering{}%
\begin{tabular}{cc}
\hline 
Model & Mean semantic distance\tabularnewline
\hline 
$\text{WBToT}(n_{d}^{y}=0.1N_{d})$ & $13.31$\tabularnewline
$\text{WBToT}(n_{d}^{y}=\sqrt{N_{d}})$ & $13.23$\tabularnewline
$\text{LDA}$ & $12.74$\tabularnewline
\hline 
\end{tabular}
\end{table}

\subsubsection{Some examples}

\begin{figure}
\begin{center}
\includegraphics[width=0.7\textwidth]{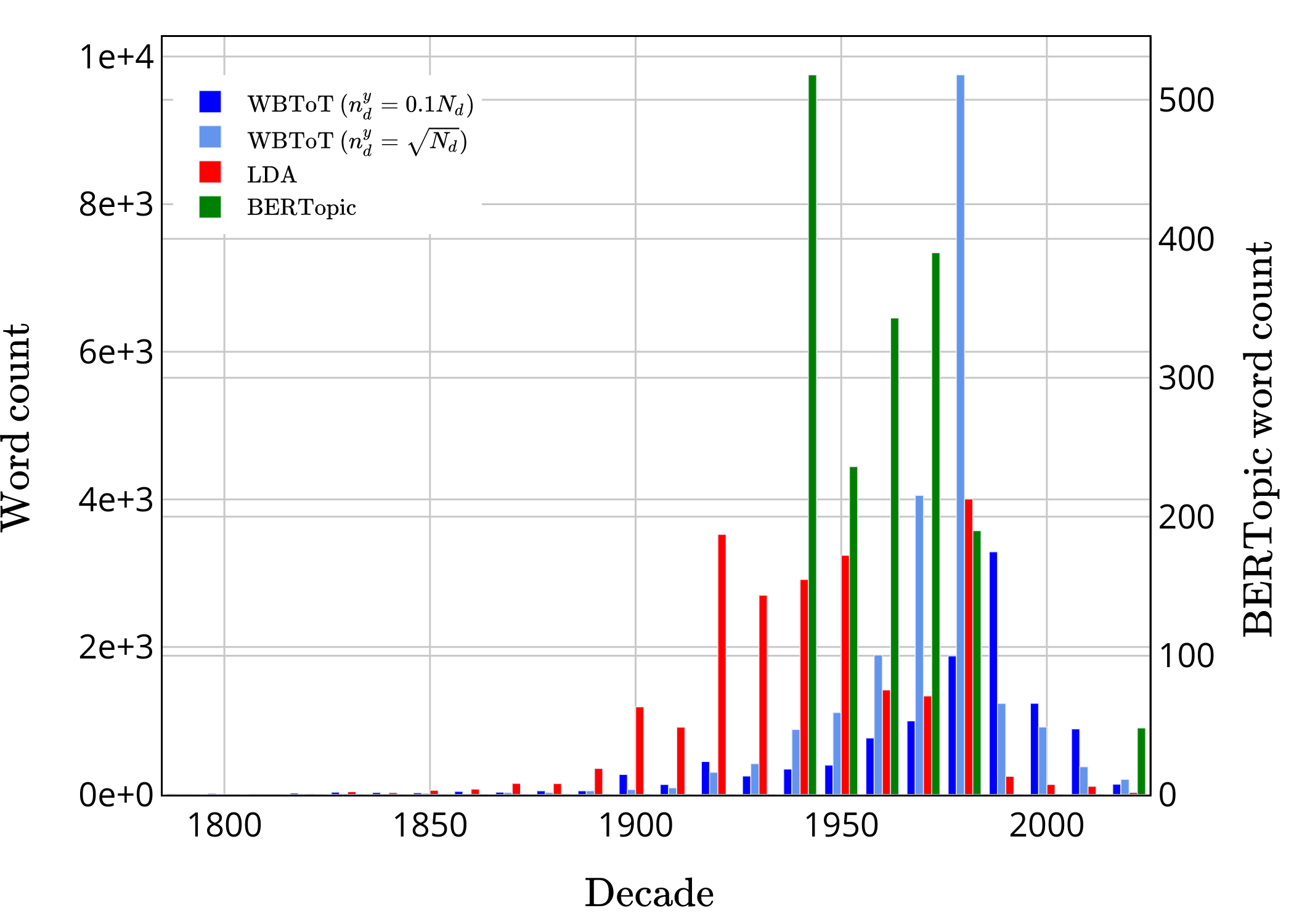}
\par\end{center}%
\begin{minipage}[c]{0.5\textwidth}%
\textbf{The Great Inflation}
\begin{flushleft}
\textbf{$\mathbf{{\text{WBToT}(n_{d}^{y}=0.1N_{d})}}$ top words}:
high, inflation, growth, increase, plan, percent, government, people,
cost, economy, american, reduce, spending, income, pay, rate, tax,
cut, budget, deficit.
\par\end{flushleft}
\begin{flushleft}
$\mathbf{{\text{\textbf{WBToT}}(n_{d}^{y}=\sqrt{N_{d}})}}$\textbf{
top words}: deficit, inflation, job, economic, cut, spending, increase,
federal, congress, economy, growth, reduce, budget, tax, program,
percent, rate, administration, high, government.
\par\end{flushleft}
\end{minipage}\hfill{}%
\begin{minipage}[c]{0.5\textwidth}%
\begin{flushleft}
\textbf{LDA top words}: private, provide, system, responsibility,
administration, local, development, policy, program, national, economic,
congress, problem, federal, continue, nation, effort, state, government,
work.
\par\end{flushleft}%
\begin{flushleft}
\textbf{BERTopic top words}: inflation, price, percent, wage, increase, growth, rise, economy, control, cost, profit, hold, high, level, job, rate, economic, food, inflationary, goods.
\par\end{flushleft}%
\end{minipage}\hfill{}%

\caption{Presence over time and 20 top words for the \textquotedblleft The
Great Inflation\textquotedblright{} topic.\label{fig:great-inflation}}
\end{figure}

\begin{figure}
\begin{center}
\includegraphics[width=0.7\textwidth]{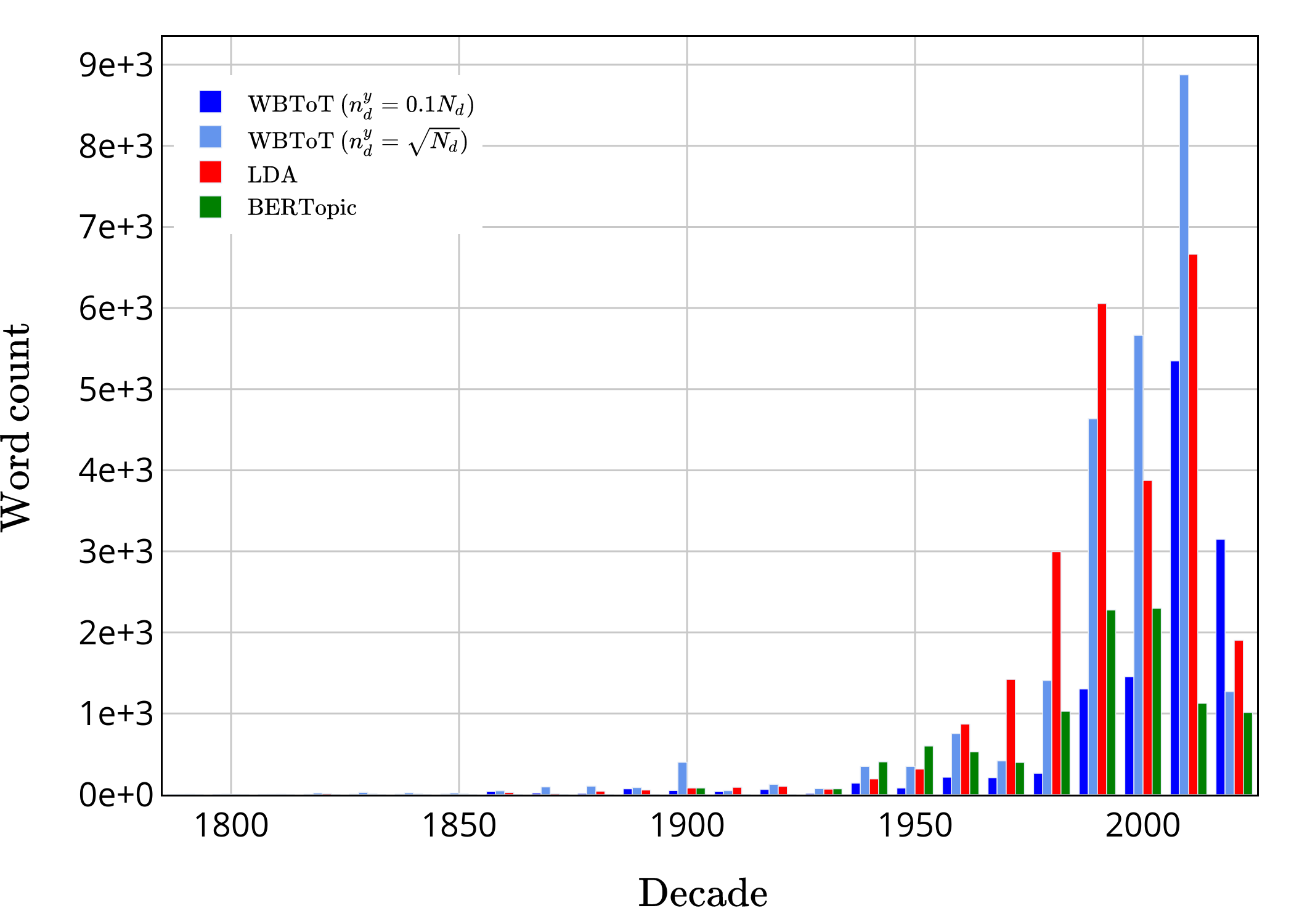}
\par\end{center}%
\begin{minipage}[c]{0.5\textwidth}%
\textbf{Health care reform}
\begin{flushleft}
\textbf{$\mathbf{{\text{WBToT}(n_{d}^{y}=0.1N_{d})}}$ top words}:
health, create, pass, time, good, economy, reform, job, country, help,
american, pay, america, plan, care, worker, business, work, people,
family.
\par\end{flushleft}
\begin{flushleft}
$\mathbf{{\text{\textbf{WBToT}}(n_{d}^{y}=\sqrt{N_{d}})}}$\textbf{
top words}: reform, college, people, america, worker, good, health,
tax, family, school, american, pay, job, child, education, business,
help, work, high, care.
\par\end{flushleft}
\end{minipage}\hfill{}%
\begin{minipage}[c]{0.5\textwidth}%
\begin{flushleft}
\textbf{LDA top words}: worker, plan, pass, deficit, congress, good,
reform, business, family, america, budget, time, economy, cut, help,
tax, people, american, job, work.
\par\end{flushleft}%
\begin{flushleft}
\textbf{BERTopic top words}: health, care, insurance, social, medical, coverage, security, medicare, family, cost, plan, drug, child, help, doctor, system, senior, reform, benefit, work.
\par\end{flushleft}%
\end{minipage}\hfill{}%

\caption{Presence over time and 20 top words for the \textquotedblleft Health
care reform\textquotedblright{} topic.\label{fig:health-care}}
\end{figure}

\begin{figure}
\begin{center}
\includegraphics[width=0.7\textwidth]{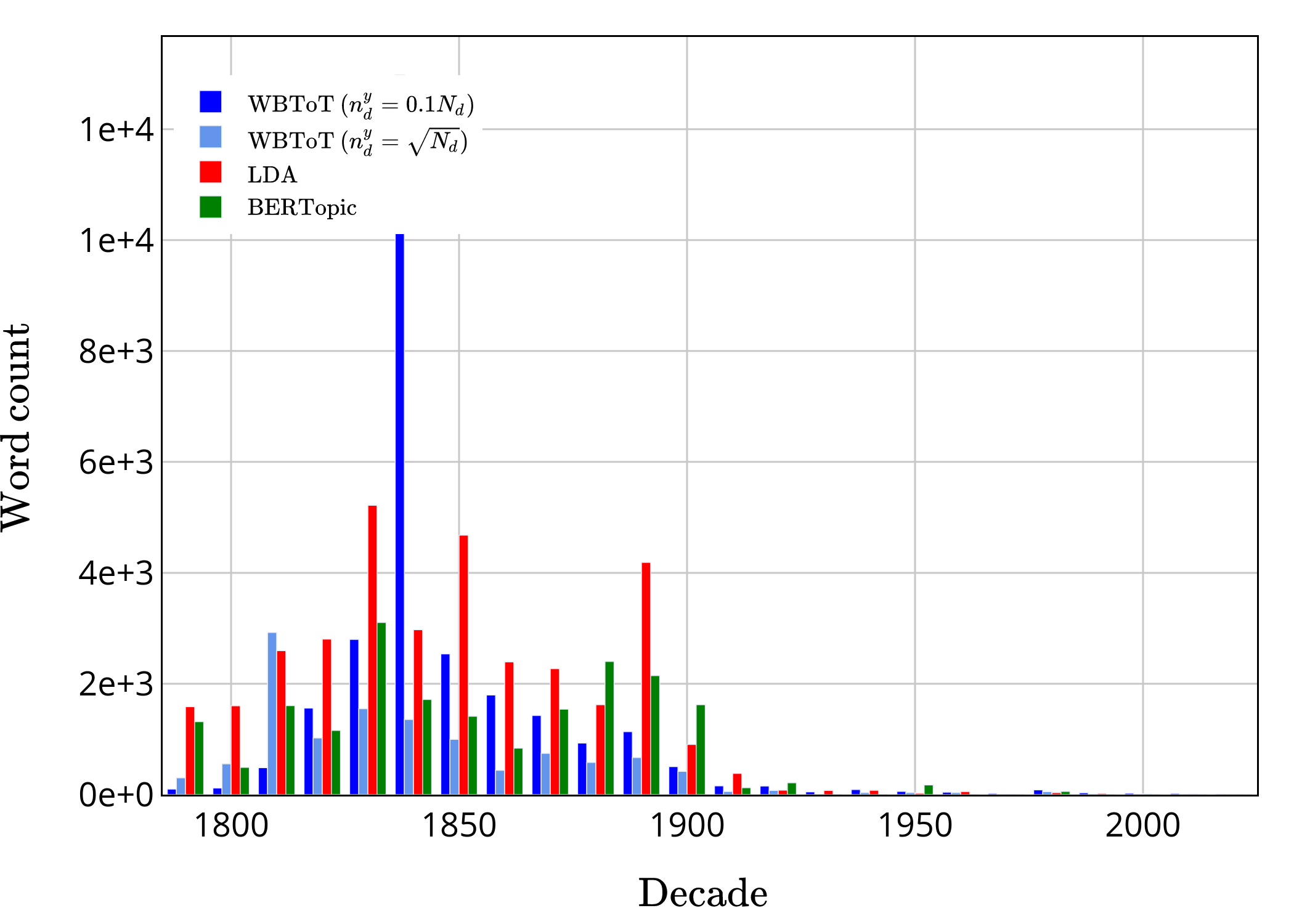}
\par\end{center}%
\begin{minipage}[c]{0.5\textwidth}%
\textbf{American Indians and conflicts with Mexico}
\begin{flushleft}
\textbf{$\mathbf{{\text{WBToT}(n_{d}^{y}=0.1N_{d})}}$ top words}:
condition, time, tribe, interest, texas, foreign, policy, people,
power, peace, indian, great, war, nation, territory, state, united,
country, government, mexico.
\par\end{flushleft}
\begin{flushleft}
$\mathbf{{\text{\textbf{WBToT}}(n_{d}^{y}=\sqrt{N_{d}})}}$\textbf{
top words}: great, white, reservation, portion, treaty, peace, civilization,
savage, settlement, policy, government, territory, country, united,
limit, state, land, indian, frontier, tribe.
\par\end{flushleft}
\end{minipage}\hfill{}%
\begin{minipage}[c]{0.5\textwidth}%
\begin{flushleft}
\textbf{LDA top words}: claim, nation, peace, spanish, hope, measure,
treaty, country, power, united, spain, relation, time, receive, citizen,
minister, government, war, state, subject.
\par\end{flushleft}%
\begin{flushleft}
\textbf{BERTopic top words}: indian, tribe, land, territory, reservation, savage, general, report, secretary, frontier, civilization, government, congress, settlement, alaska, mississippi, white, allotment, state, united.
\par\end{flushleft}%
\end{minipage}\hfill{}%

\caption{Presence over time and 20 top words for the \textquotedblleft American
Indians and conflicts with Mexico\textquotedblright{} topic.\label{fig:american-indians-mexico}}
\end{figure}

\begin{figure}
\begin{center}
\includegraphics[width=0.7\textwidth]{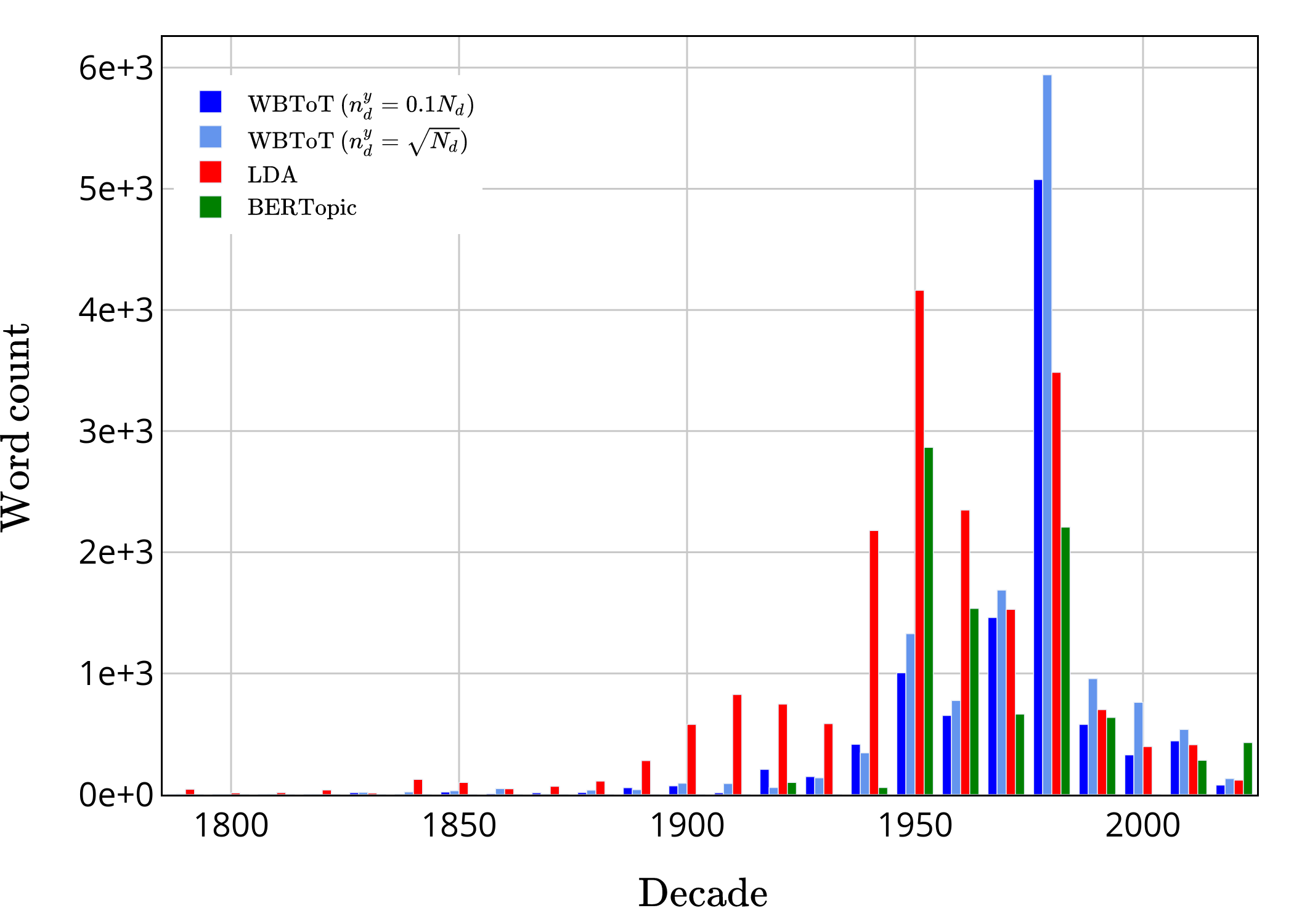}
\par\end{center}%
\begin{minipage}[c]{0.5\textwidth}%
\textbf{Cold War}
\begin{flushleft}
\textbf{$\mathbf{{\text{WBToT}(n_{d}^{y}=0.1N_{d})}}$ top words}:
strong, arm, union, united, force, strategic, agreement, continue,
nation, security, nuclear, policy, soviet, military, trade, america,
economic, international, peace, defense.
\par\end{flushleft}
\begin{flushleft}
$\mathbf{{\text{\textbf{WBToT}}(n_{d}^{y}=\sqrt{N_{d}})}}$\textbf{
top words}: rights, east, war, union, nuclear, strong, soviet, america,
force, future, continue, people, nation, peace, security, military,
human, state, free, freedom.
\par\end{flushleft}
\end{minipage}\hfill{}%
\begin{minipage}[c]{0.5\textwidth}%
\begin{flushleft}
\textbf{LDA top words}: defense, effort, aggression, europe, continue,
policy, help, strength, agreement, force, economic, international,
united, free, soviet, peace, nation, military, war, security.
\par\end{flushleft}%
\begin{flushleft}
\textbf{BERTopic top words}: soviet, nuclear, communist, free, union, defense, military, force, threat, arm, ally, europe, korea, weapon, aggression, strategic, nation, agreement, security, peace.
\par\end{flushleft}%
\end{minipage}\hfill{}%

\caption{Presence over time and 20 top words for the \textquotedblleft Cold War
\textquotedblright{} topic.\label{fig:cold-war}}
\end{figure}

In order to get an intuition of how well events are captured across the different models in this experiment (LDA, BERTopic, and the two WBToT models), let us sample some topics that show a clear link
with historical events. For each topic in one model, we will identify the closest topics in the other models. The definition of "closest topics" is according to the symmetric
KL divergence introduced in Eq. (\ref{eq:sym_kl_div}) for LDA and WBToT, and to word overlaps in the case of BERTopic (due to its problem with Kullback-Leibler divergence, as noted in the previous section). In each case, we will
plot the presence of each topic over time (represented as a histogram
over decades) and the top 20 words that characterize them. Results
are shown in Fig. \ref{fig:great-inflation} - \ref{fig:cold-war}.
We use two tones of blue to represent the two WBToT models, red for LDA and green for BERTopic.

In Fig. \ref{fig:great-inflation}, we observe a set of topics associated
with inflation. In particular, it seems that the WBToT models have captured
the event known as ``The Great Inflation'', a historical period
of high inflation that peaked around the 1980s. The top words in these
two models make clear reference to concepts such as ``high'', ``inflation'',
``increase'', ``percent'', and ``cost''. None of these words
appear in the LDA model, which seems to deal with more general economic
issues. Notice that the topic presence over time is almost flat during
most part of the 20th century for LDA, while histograms for both WBToTs
models exhibit clear peaks around the 80s. To us, this is an indication
that the WBToT models have captured a concrete period of American
history characterized by high inflation, while the LDA is modeling
economic issues in general. BERTopic correctly forms a topic about inflation,
with a set of top words that resemble those of WBToT. However, this topic is not sufficiently
populated in BERTopic: we have been forced to use two different y-axes in Fig. \ref{fig:great-inflation} (one specific for BERTopic) due to the small number of documents that have been assigned to this cluster. The peak is not clear either, since the topic distribution spans the entire second half of the 20th century.

In Fig. \ref{fig:health-care} a set of topics associated to the
healthcare system reform has been discovered. In this case, LDA performs
pretty well in comparison with both WBToT. However, the WBToT model
with $n_{d}^{\left(y\right)}=0.1N_{d}$ seems to capture better than
any other model the debate around the Affordable Care Act (colloquially
known as Obamacare), a polemic reform that was signed into law in
2010 and came into force during the rest of the decade. This WBToT
model shows a clear peak in the 2010s with some important repercussions
also in the 2020s and residual presence before that. The WBToT model
with $n_{d}^{\left(y\right)}=\sqrt{N_{d}}$ actually shows a higher
peak during the 2010s, although its presence is also noticeable in
the preceding decades, so it has not been able to capture the event
as precisely as the first one. The LDA model, on the other hand, shows
a more dispersed histogram with two different peaks (one in the 2010s
and the other in the 1990s) and tends to capture a broader vocabulary
related to other types of reforms (notice that the keywords ``health''
and ``care'', which are present in both WBToT models, do not appear
in the LDA top words). BERTopic doesn't exhibit the expected presence over time, with a peak around the 1990s. The top words are pretty cohesive and clearly describe a topic about healthcare, but its distribution over time seems generic.

In Fig. \ref{fig:american-indians-mexico} we see a set of topics
that captures some of the main conflicts that the incipient US faced
on North American territory during the 19th century: in particular,
conflicts with American Indians and with Mexico. It is interesting
to note that the WBToT model with $n_{d}^{\left(y\right)}=0.1N_{d}$
is more focused on the Mexican-American conflicts, with a clear peak
in the 1840s (the Mexican-American war took place between April 1846
and February 1848, so it seems to capture this event in a very precise
manner), while the WBToT model with $n_{d}^{\left(y\right)}=\sqrt{N_{d}}$
put more emphasis on American-Indian conflicts, with a peak in a much
earlier period, around the 1810s. The LDA, however, exhibits a more
disperse distribution that spans the entire 19th century (with three
different peaks at different times), and includes other unrelated
events such as the Spanish-American war in the final years of the
century. In this case, BERTopic also exhibits a dispersed distribution that spans the entire 19th century, although its top words don't mix different topics.

In Fig. \ref{fig:cold-war}, a set of topics associated with the Cold
War has been represented. Both WBToT models exhibit a clear peak in
the second half of the 20th century (with the 1980s as the more populated
decade) and vanishes quickly in the first half of the century. The
LDA model also exhibits a strong presence during the second half of
the century, but there is a noticeable presence during the first half
of the century as well (probably due to the Second World War) and some residual presence even before the
20th century, which is clearly out of the period spanned by the Cold
War. By looking at the top words of each model, we can see that LDA
seems to exhibit a more general vocabulary that could be used to describe
other types of wars and military tensions: in particular, it is interesting
to note that both WBToT models feature ``nuclear'' as a top word
(a concept that defined one of the key singularities of the Cold War
with respect to other conflicts), while LDA lacks that word. BERTopic exhibits an excellent semantic representation of the topic, with very precise top words. However, its distribution over time exhibits a relative spike in the decades of the 2010s and the 2020s, long after the end of the conflict.

\subsection{Coherence\label{subsec:wbtot_vs_btot}}

The main motivation to introduce the WBToT model as an extension of the BToT model was to balance the
excessive importance that BToT places on timestamps. As already stated
in Section \ref{subsec:WBToT}, in BToT the same timestamp, $t_{d}$,
is generated for each word of document $d$. WBToT solves this problem
by introducing an additional latent variable, $y$, that determines
the number of times that the same timestamp is observed per document.
This feature allows WBToT to generate topics that are more semantically
coherent, instead of focusing almost exclusively on modeling publication
timestamps as BToT does.

Topic coherence measures are typically used to assess topic's semantic quality and interpretability. We employ the framework described in \citet{Roder2015} and adopt the coherence definition that exhibits the best overall performance in their experimental setting, $C_V$. This is a composition of a one-set segmentation of top words, a boolean sliding window, and an indirect confirmation measure that combines normalized pointwise mutual information and cosine similarity. In order to compute coherence,
a list of words that identify each topic must be provided. The standard
option is to rank words according to $\beta_{kw}$ for each topic
$k$ and select a certain number of top words (we will select the
10 top words). Other scoring functions such as $r_{kw}^{\text{log}}$,
\begin{equation}
r_{kw}^{\text{log}}\coloneqq\beta_{kw}\left(\log\beta_{kw}-\left\langle \log\beta_{kw}\right\rangle _{k}\right)\label{eq:log_rank}
\end{equation}
can be used as well. $r_{kw}^{\text{log}}$ is a very common scoring function whose merit lies in demoting very frequent words, so the  ``essence''
of the topic is better appreciated. In this section, we will compute coherence by scoring topic words both with $\beta_{kw}$and $r_{kw}^{\text{log}}$.

\begin{figure}
\subfloat[Using the $\beta_{kw}$ rank\label{fig:beta-rank-coherence}]{\includegraphics[width=0.47\textwidth]{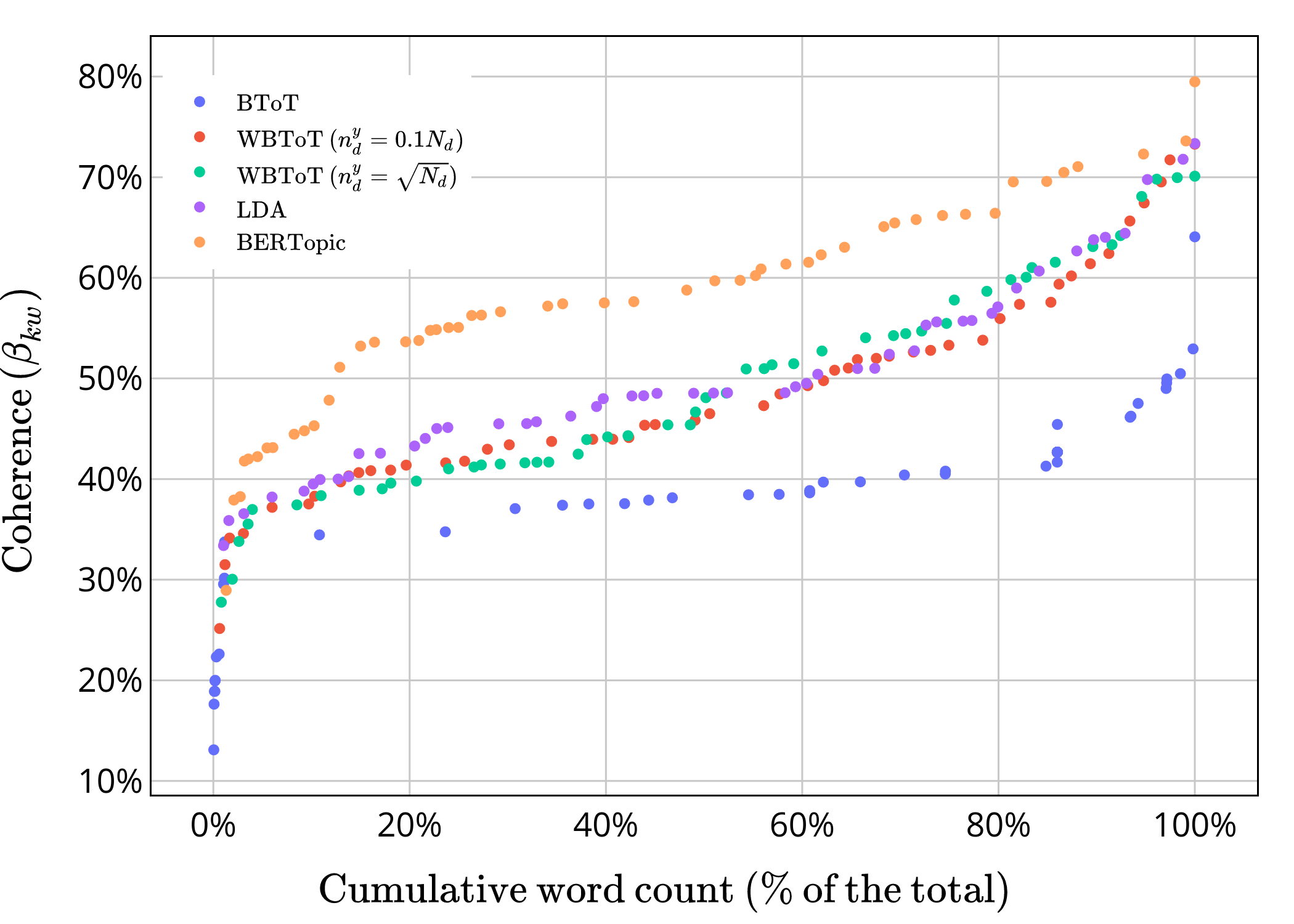}

}\hfill{}\subfloat[Using the $r_{kw}^{\log}$ ranking function\label{fig:r-rank-coherence}]{\includegraphics[width=0.47\textwidth]{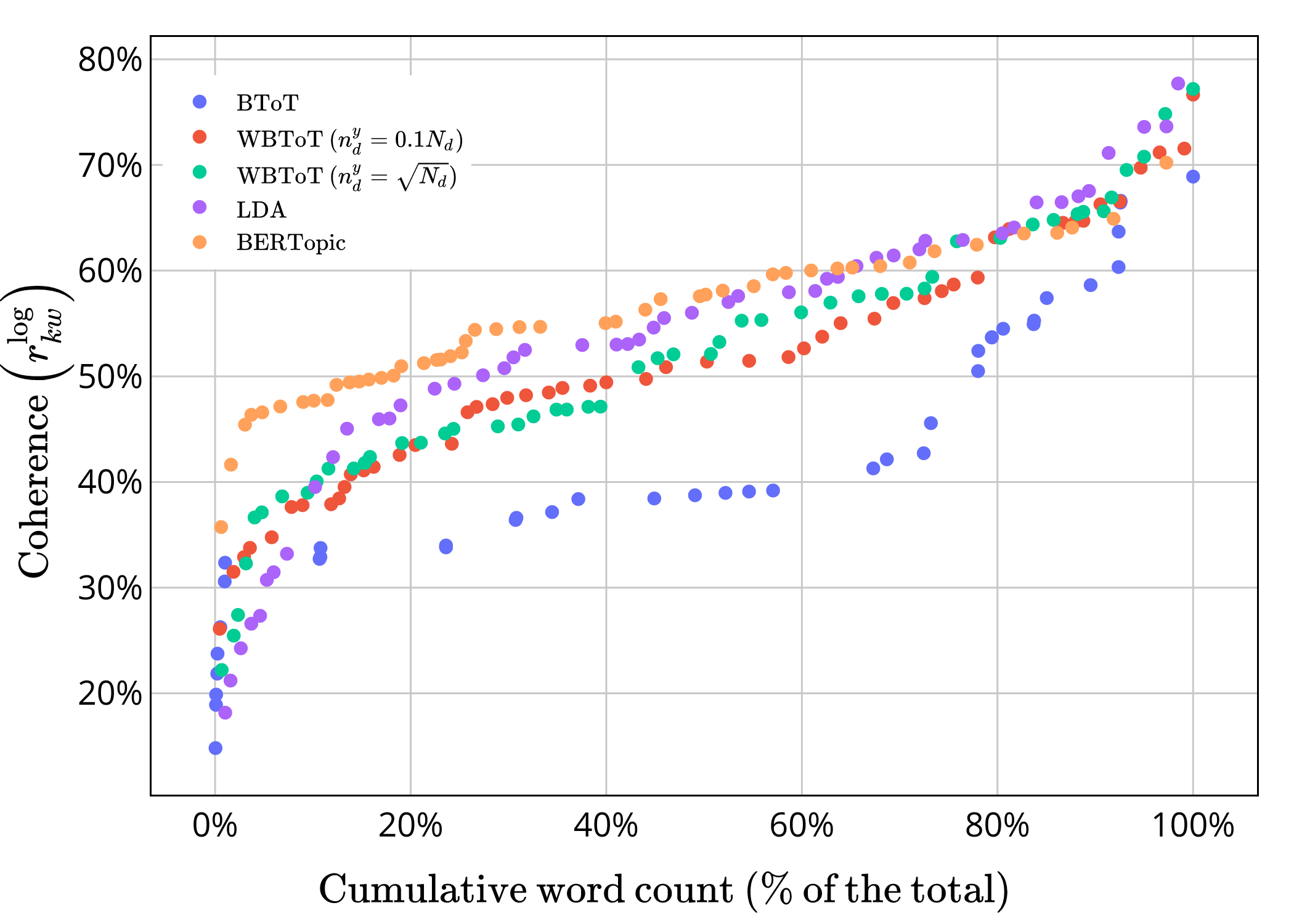}}

\caption{Model coherence as a function of cumulative word count}
\label{fig:wbtot_vs_btot}
\end{figure}

For this comparison, we compute the coherence of each of the 50 topics
found in the SOTU dataset (see Section \ref{subsec:sotu_dataset})
for LDA, BERTopic, BToT, and the two WBToT models. Since mean values might
be affected by outliers, it is not a good idea to simply measure the
average coherence for each collection of topics. Instead, it is more
interesting to measure the number of tokens assigned to topics with
less (or equal) coherence than a certain value. To achieve this, we
will sort topics from less to more coherent (for each topic model).
Then we compute the number of tokens assigned to each of these topics
(that is, $\sum_{d}\theta_{dk}n_{dw}$) and we compute the cumulative
sum of tokens for the coherence-sorted list of topics. Results are
presented in Fig. \ref{fig:wbtot_vs_btot}, where coherence has
been computed using $\beta_{kw}$ for Fig. \ref{fig:beta-rank-coherence}
and $r_{kw}^{\text{log}}$ for Fig. \ref{fig:r-rank-coherence}. 

It is clear that the BToT model exhibits coherence values that consistently
rank below those of the two WBToT models, especially in the midsection
of the graphs. In fact, we discovered that 10 out of 50 topics in
BToT were assigned a flat word probability (i.e. $\beta_{kw}\backsimeq\beta_{k0}$),
which implies that they do not convey any semantic meaning at all.
This result supports the introduction of an additional latent variable
in WBToT, and therefore can be seen as an ablation study that proves the importance of balancing the different modalities in the log-likelihood. 

Both versions of the WBToT model (depending on the $n_{d}^{\left(y\right)}$
parameter) perform similarly in terms of coherence, and it is not clear
if one of them is better than the other.

We also observe that the introduction of a time modality in WBToT does not 
degrade much the coherence with respect to LDA. In both figures, LDA coherence is similar
to that of these two models, which implies that they preserve the quality of semantic meaning.

BERTopic coherence ranks above all the studied models when top words are ordered according to $\beta_{kw}$,
but this is not so clear when we use $r_{kw}^{\text{log}}$ instead. Furthermore, it must be
noticed that BERTopic automatically removes documents that are classified as outliers, which
naturally improves the coherence of topics. In WBToT, we could also remove ill-represented documents as a final step, but since coherence is not our priority, we leave this for future research.

\subsection{Stability of WBToT\label{subsec:wbtot_online_version}}

Standard ToT already captured events better than LDA \citep{Wang2006}. Nevertheless, the main practical advantage of using a fully Bayesian model is the
fact that instabilities are prevented when a topic
is poorly represented in a mini-batch or exhibits a very peaked timestamp
structure (recall Eqs. (\ref{eq:rho_k0_eq_1}) - (\ref{eq:rho_k0_eq_2}) and the subsequent discussion, which mathematically proves the instability of ToT online updates). Therefore, with WBToT one can make use of a stable online optimization approach which is not present in standard ToT.

\begin{figure}
\centering{}\includegraphics[width=0.5\textwidth]{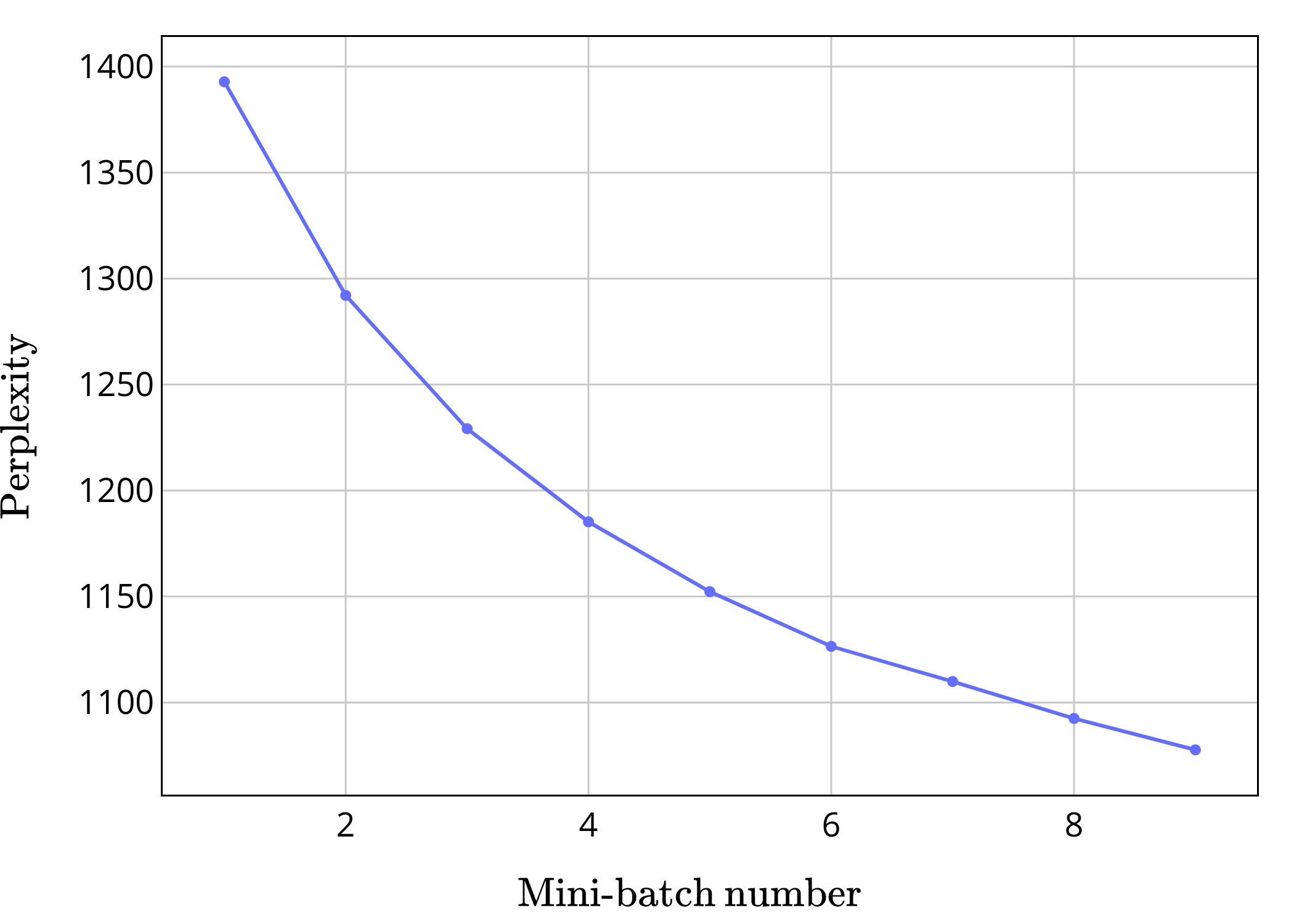}\caption{Perplexity evolution with the number of analyzed mini-batches.\label{fig:wbtot_online_perp}}
\end{figure}

To illustrate the stability of the online algorithm in WBToT, we will consider
the experiment on the large-scale COVID-19 Twitter dataset described
in Section \ref{subsec:covid_dataset}. To begin with, we prove that perplexity
on a held-out dataset decreases as successive mini-batches are analyzed.
To do this, we evaluate perplexity on the test set (which is composed
of 1 million random tweets) each time the model optimizes its parameters
after receiving a new mini-batch. As we mentioned before, we have
considered 9 mini-batches of 1 million tweets each. Results are presented
in Fig. \ref{fig:wbtot_online_perp}, where we have represented
perplexity on the test set as a function of the number of mini-batches
analyzed. Notice that only one iteration of the EM algorithm has been
performed, but the perplexity seems to approach convergence quite
fast.

\begin{figure}
\subfloat[After mini-batch 1]{\includegraphics[width=0.32\textwidth]{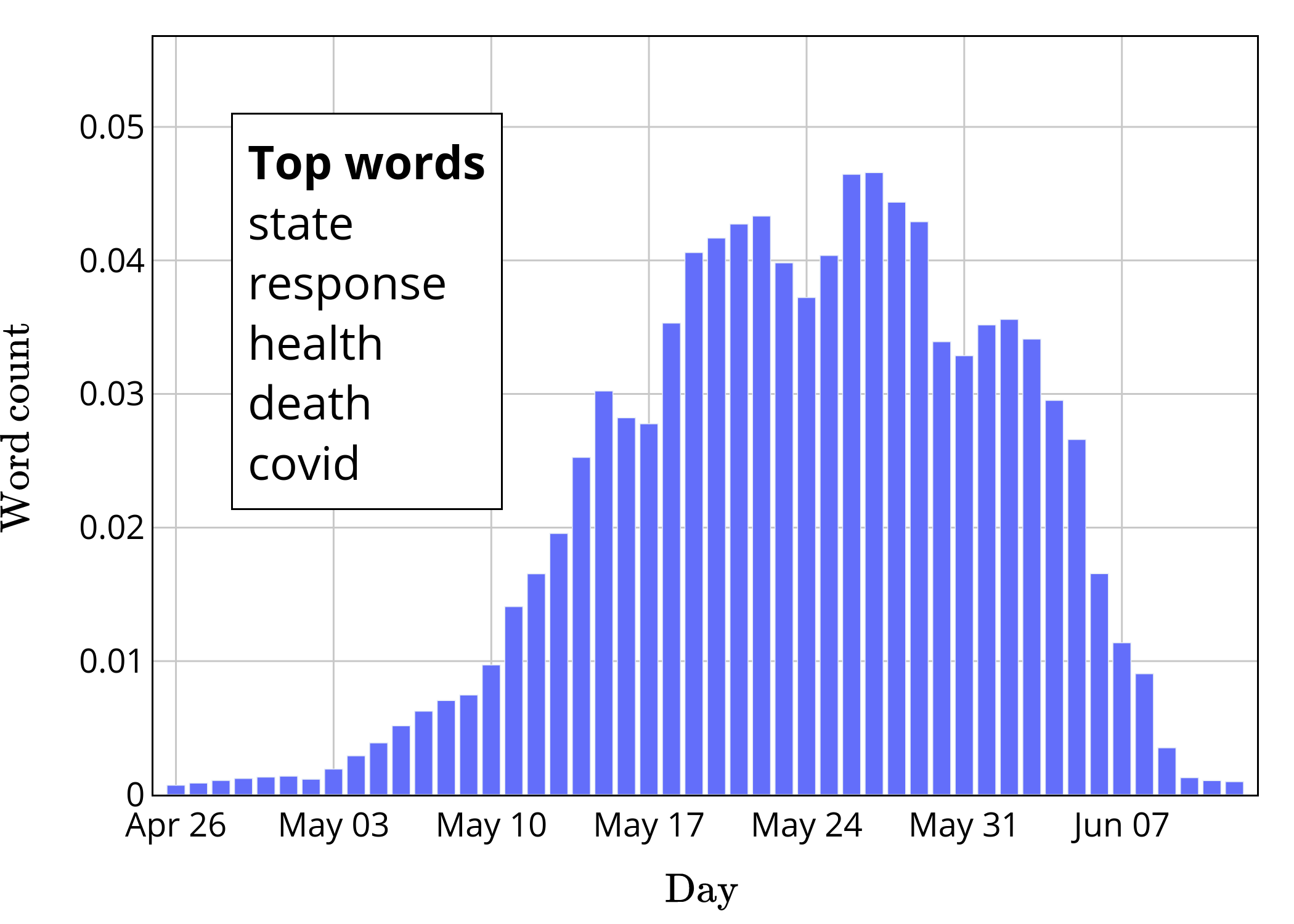}

}\hfill{}\subfloat[After mini-batch 5]{\includegraphics[width=0.32\textwidth]{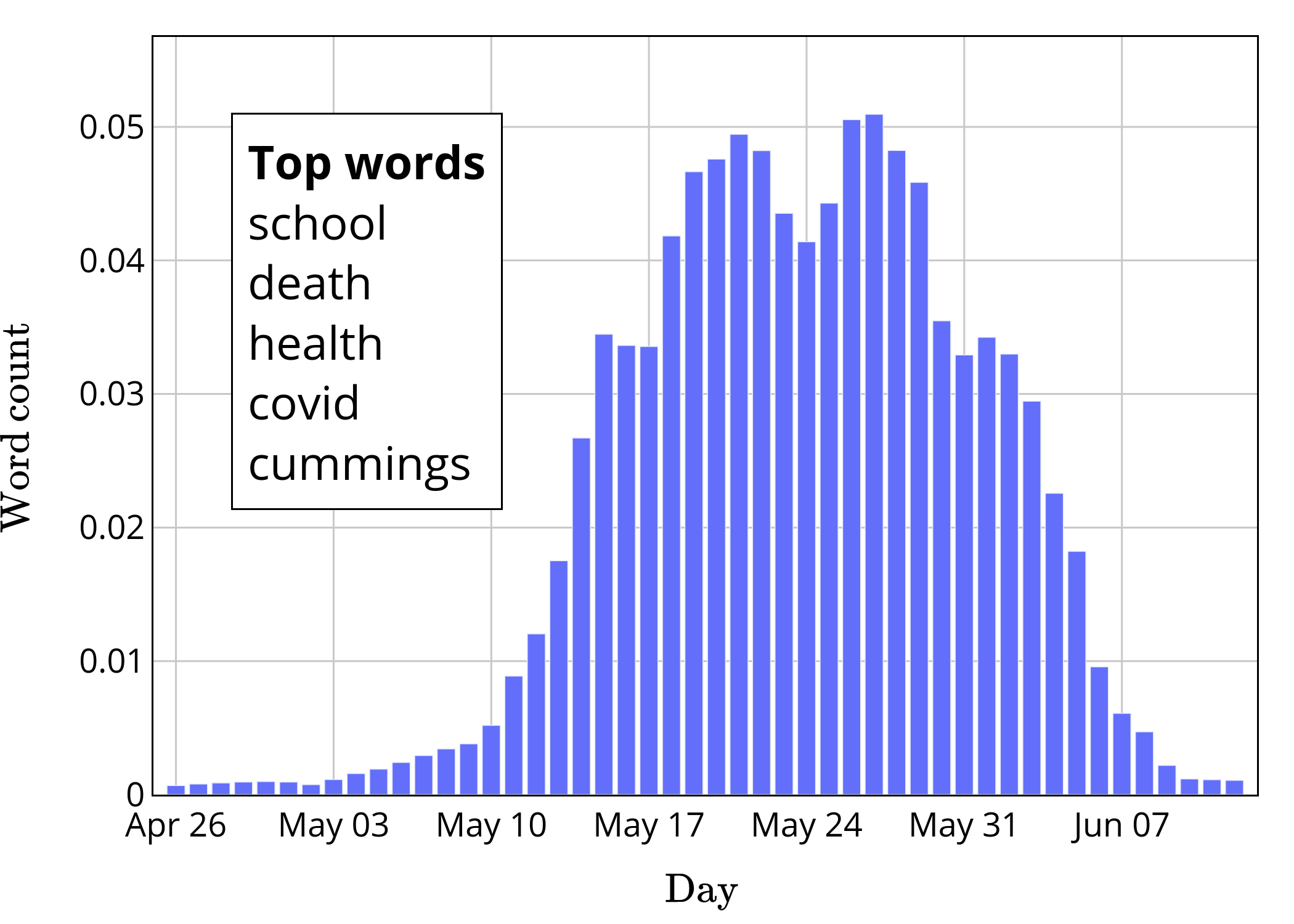}

}\hfill{}\subfloat[After last mini-batch]{\includegraphics[width=0.32\textwidth]{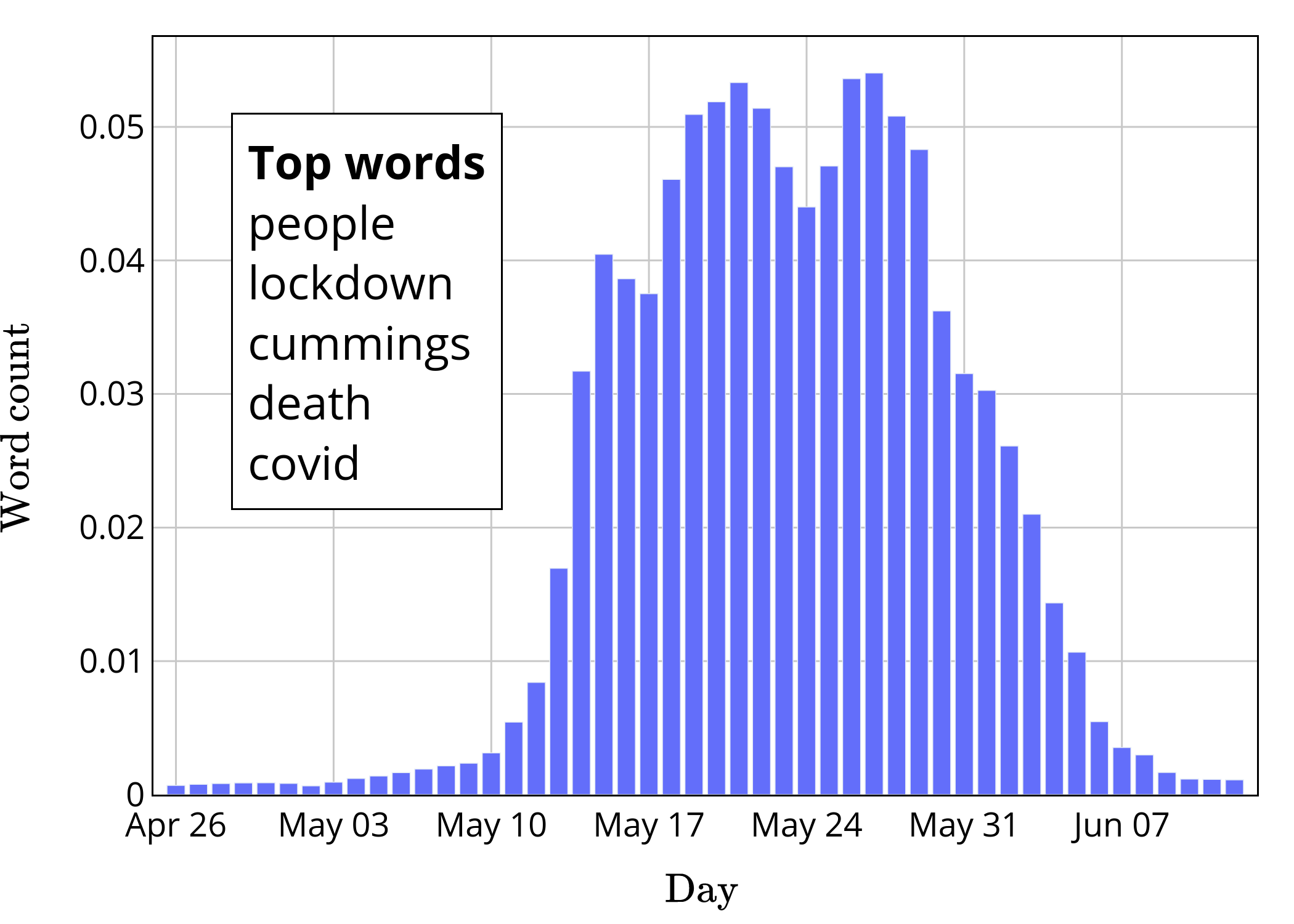}

}

\caption{Evolution of topic regarding COVID-19 lockdowns.\label{fig:wbtot_online_topic_id_0}}
\end{figure}

\begin{figure}
\subfloat[After mini-batch 1]{\includegraphics[width=0.32\textwidth]{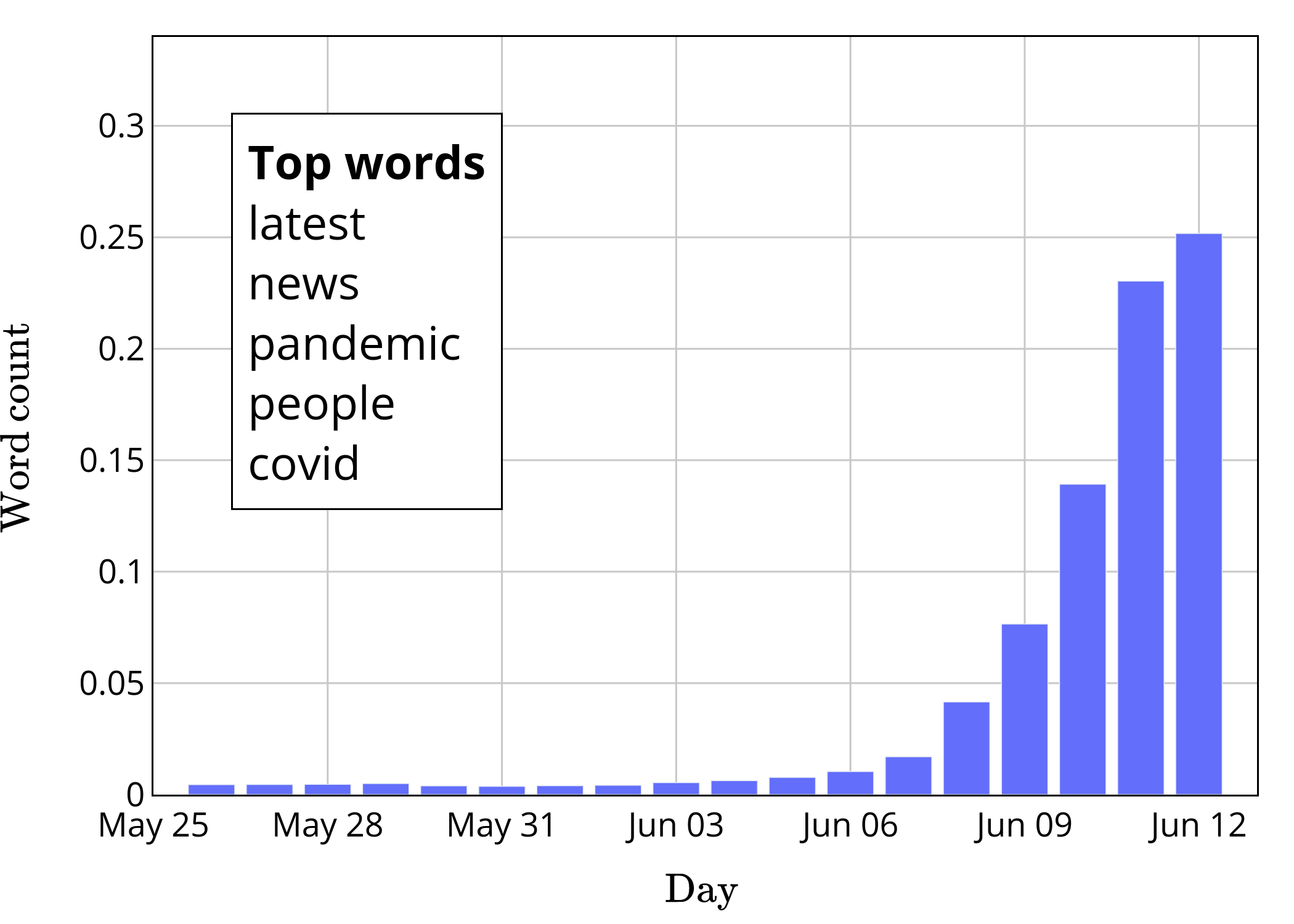}

}\hfill{}\subfloat[After mini-batch 5]{\includegraphics[width=0.32\textwidth]{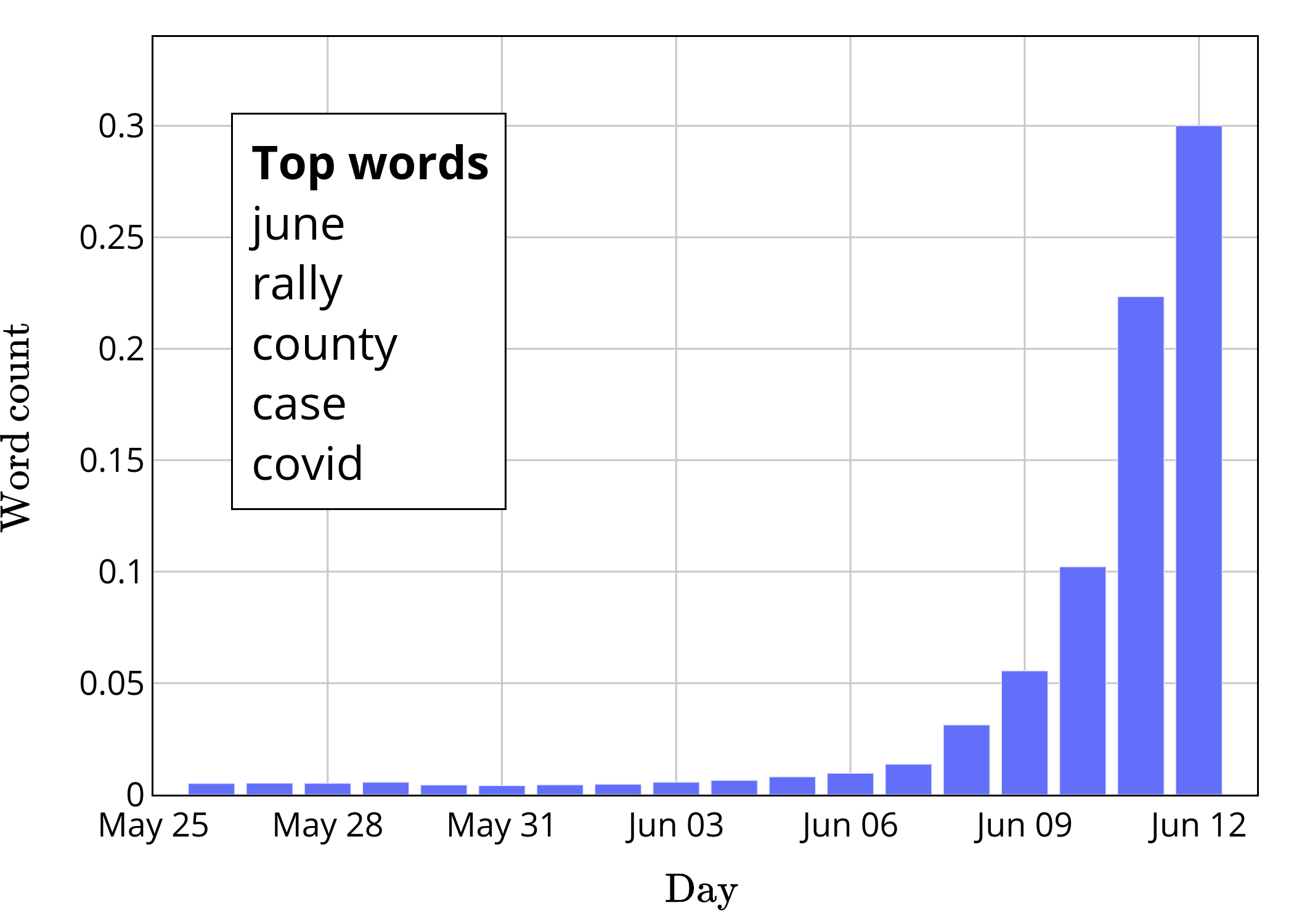}

}\hfill{}\subfloat[After last mini-batch]{\includegraphics[width=0.32\textwidth]{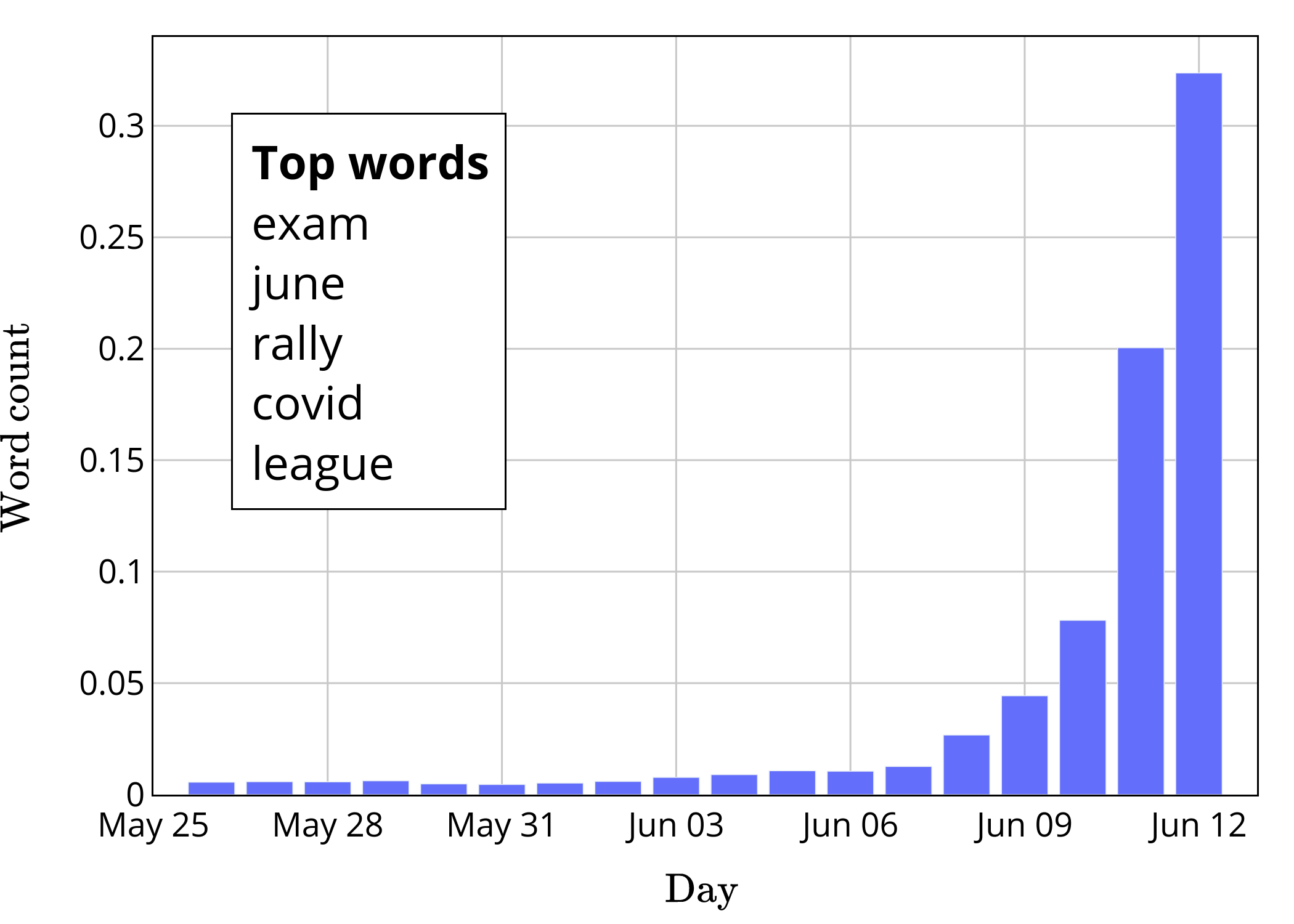}

}

\caption{Evolution of topic regarding June exams and Tulsa rally.\label{fig:wbtot_online_topic_id_2}}
\end{figure}

Next, as an example, we present the evolution of a couple of topics
discovered by the model. In Fig. \ref{fig:wbtot_online_topic_id_0}
- \ref{fig:wbtot_online_topic_id_2} we plot the normalized presence of each
topic over time and the top five words that describe them at three
different stages of training: after analyzing the first mini-batch
(left subfigures), after four mini-batches (central subfigures) and
after all the mini-batches (right subfigures). Since the vocabulary
is very similar for all the topics, we have ranked the top words according to $r_{kw}^{\text{log}}$, as defined in Eq. (\ref{eq:log_rank}), instead of $\beta_{kw}$ to increase the readability of the topics. 

In Fig. \ref{fig:wbtot_online_topic_id_0} we observe a topic that
mainly deals with lockdowns and deaths associated to COVID-19. It
starts as a relatively flat topic that comprises the entire month
of May dealing with generic governments' responses to COVID-19. However,
as training progresses, it becomes more and more peaked, and the vocabulary gravitates towards specific events that occurred at that time.

In Fig. \ref{fig:wbtot_online_topic_id_2} we observe a topic with
a very sharp peak at the end of the period considered here (middle
of June 2020). The top words after the first mini-batch are quite generic,
but they become clearer as training progresses: the topic focuses
on events that defined the month of June, such as the management of
exams and political rallies for the first time since the start of the pandemic.

We tried to implement an online algorithm for standard ToT in an attempt to reproduce these results, but instabilities led to numerical errors. This result confirms the expected behavior that motivated our search for fully Bayesian models.

\section{Conclusions\label{sec:Conclusions}}

In this work, we reformulated the ToT model to make it fully Bayesian (BToT) by introducing a prior for the Beta distribution. In order to deal with the different scales of the two modes of the log-likelihood (words and timestamps), a modified version of this model, WBToT, was proposed. In WBToT, the number of timestamp observations is reduced by the introduction of an additional latent topic assignment variable.

Our experiments prove that WBToT captures events better than
LDA while maintaining a similar level of coherence. Our experiments also illustrated
that we can make use of an online optimization algorithm in WBToT, a feature that is not present in standard ToT due to stability issues (originating from the lack
of a prior for the Beta distribution).

WBToT also exhibits better event modelization than embedding-based methods such as BERTopic. However, the superior coherence shown by BERTopic opens up an interesting line of research: can BERTopic be extended to take timestamps into account and better capture events? We will address this question in future works.

Notice that we have worked with news and tweets separately. In future works, we would like to investigate how our model performs on a collection of multiple and heterogeneous sources (news, tweets, RSS feeds...) to test if it's capable of recognizing events across different streams of data \citep{Mele2019}.

The WBToT model solves the main shortcoming of the popular ToT model. The introduction
of a robust online optimization method allows
using this model for practical applications that were problematic
for standard ToT (see, for example, \citealt{Al-Ghossein2018}). Furthermore,
we believe that our approach could also be integrated into any of the
extensions of ToT described in Section \ref{sec:Related-work}, such as
the AToT \citep{Xu2014} and the HTMOT \citep{Poumay2021} models.

\section*{Acknowledgment}

This work was supported by the Spanish Centre for the Development of Industrial Technology (CDTI) under grant number IDI-20190525. The work of Julio Gonzalo was partially supported by the Spanish Ministry of Science and Innovation through Project FairTransNLP under Grant PID2021-124361OB-C32. Any opinions, findings and conclusions or recommendations expressed in this material are those of the authors and do not necessarily
reflect those of the sponsor.

\appendices{}

\section{Description of LDA\label{sec:Description-of-LDA}}

The main references for the results in this appendix are \citet{Blei2003,Hoffman2010}.
The generative process for LDA can be described as:
\begin{enumerate}
\item Draw $K$ multinomial parameters $\boldsymbol{\beta_{k}}$ from $K$ Dirichlet distributions
with parameters $\boldsymbol{\eta_{k}}$.
\item For each document $d$, draw a set of multinomial parameters $\boldsymbol{\theta_{d}}$ from a symmetric
Dirichlet distribution with parameters $\boldsymbol{\alpha}$. Then, for each word
index $i\in\{1,\ldots, N_{d}\}$ in document $d$:
\begin{enumerate}
\item Draw a topic $z_{di}$ from a multinomial with parameters $\boldsymbol{\theta_{d}}$;
\item Draw a word $w_{di}$ from a multinomial with parameters $\boldsymbol{\beta_{z_{di}}}$.
\end{enumerate}
\end{enumerate}
The corresponding log-likelihood has the
form:

\begin{equation}
\begin{aligned}\mathcal{L}^{\text{LDA}}= & \sum_{d=1}^{D}\left\{ \log\underbrace{p\left(\boldsymbol{\theta_{d}}|\boldsymbol{\alpha}\right)}_{\text{Dirichlet}}+\sum_{i=1}^{N_{d}}\left[\log\underbrace{p\left(z_{di}|\boldsymbol{\theta_{d}}\right)}_{\textrm{Multinomial}}+\log\underbrace{p\left(w_{di}|\boldsymbol{\beta_{z_{di}}}\right)}_{\text{Multinomial}}\right]\right\} +\sum_{k=1}^{K}\log\underbrace{p\left(\boldsymbol{\beta_{k}}|\boldsymbol{\eta_{k}}\right)}_{\textrm{Dirichlet}}.\end{aligned}
\label{eq:LDAlogLikehood}
\end{equation}

In the variational inference approach, the true posterior distribution
is approximated by a fully factorized distribution $q^{\text{LDA}}$,
indexed by the variational parameters $\phi_{dwk}$, $\gamma_{dk}$
and $\lambda_{kw}$, of the form:
\begin{equation}
\begin{aligned}q^{\text{LDA}}=\prod_{k=1}^{K}\underbrace{q\left(\boldsymbol{\beta_{k}}|\boldsymbol{\lambda_{k}}\right)}_{\text{Dirichlet }} & \prod_{d=1}^{D}\left[\underbrace{q\left(\boldsymbol{\theta_{d}}|\boldsymbol{\gamma_{d}}\right)}_{\text{Dirichlet}}\prod_{i=1}^{N_{d}}\underbrace{q\left(z_{di}|\boldsymbol{\phi_{dw_{di}}}\right)}_{\text{Multinomial}}\right].\end{aligned}
\label{eq:LDAVariationalAnsatz}
\end{equation}
The evidence lower bound (ELBO) is 

\begin{equation}
\mathcal{L^{\text{LDA-ELBO}}}=\left\langle \mathcal{L^{\text{LDA}}}-\log q^{\text{LDA}}\right\rangle _{q^{\text{LDA}}}\label{eq:lda_elbo}
\end{equation}
and its maximization leads to the standard VB LDA equations for the
variational parameters
\begin{equation}
\phi_{dwk}\propto\exp\left[\left\langle \log\beta_{kw}\right\rangle _{q^{\text{LDA}}}+\left\langle \log\theta_{dk}\right\rangle _{q^{\text{LDA}}}\right],\label{eq:lda_phi_equation}
\end{equation}
\begin{equation}
\gamma_{dk}=\alpha_{k}+\sum_{w}n_{dw}\phi_{dwk},\label{eq:lda_gamma_equation}
\end{equation}
\begin{equation}
\begin{aligned}\lambda_{kw}= & \eta_{kw}+\sum_{d}n_{dw}\phi_{dwk}.\end{aligned}
\label{eq:lda_lambda_equation}
\end{equation}
The Dirichlet averages appearing in Eq. (\ref{eq:lda_phi_equation}) are:

\begin{equation}
\left\langle \log\theta_{dk}\right\rangle _{q^{\text{LDA}}}=\Psi\left(\gamma_{dk}\right)-\Psi\left(\sum_{k'}\gamma_{dk'}\right),\label{eq:theta_q_mean}
\end{equation}
\begin{equation}
\left\langle \log\beta_{kw}\right\rangle _{q^{\text{LDA}}}=\Psi\left(\lambda_{kw}\right)-\Psi\left(\sum_{w'}\lambda_{kw'}\right),\label{beta_q_mean}
\end{equation}
where $\Psi\left(z\right):=d\log\Gamma\left(z\right)/dz$ is the digamma
function.

In case of an online optimization, Eq. (\ref{eq:lda_lambda_equation}) is
modified to
\begin{equation}
\lambda_{kw}\left(t+1\right)=\left(1-\rho_{t}\right)\lambda_{kw}\left(t\right)+\rho_{t}\left(\eta_{kw}+\frac{D}{S}\sum_{d\in\text{MB}}n_{dw}\phi_{dwk}\right).\label{eq:lda_lambda_equation_online}
\end{equation}
where $\text{MB}$ is a mini-batch of $S$ documents. Both batch and
online methods are summarized in Algorithm \ref{alg:BatchLDA} - \ref{alg:OnlineLDA}.
\begin{algorithm}
\caption{\label{alg:BatchLDA} Batch variational LDA}
\begin{algorithmic}
\STATE Initialize variational parameters $\gamma_{dk},\lambda_{kw}$
\REPEAT
\STATE \emph{E-step:}
\FOR{$d=1$ to $D$}
\REPEAT
\STATE Set $\phi_{dwk},\gamma_{dk}$ from Eq. (\ref{eq:lda_phi_equation}) and Eq. (\ref{eq:lda_gamma_equation}).
\UNTIL $\gamma_{dk}$ converged.
\ENDFOR
\STATE \emph{M-step:}
\STATE Set $\lambda_{kw}$ from Eq. (\ref{eq:lda_lambda_equation}).
 \UNTIL $\mathcal{L}$ converged.
\end{algorithmic}
\end{algorithm}

\begin{algorithm}
\caption{\label{alg:OnlineLDA} Online variational LDA}
\begin{algorithmic}

\STATE Set $t=0$.
\STATE Initialize variational parameters $\lambda_{kw}(t)$.
\REPEAT
\STATE Set $t \gets t+1$.
\STATE Set $\rho_{t}:=\left(t+\tau\right)^{-\kappa}$.
\STATE \emph{E-step:}
\FOR{$d=1$ to $S$}
\STATE Initialize variational parameters $\gamma_{dk}$.
\REPEAT 
\STATE Set $\phi_{dwk},\gamma_{dk}$ from Eq. (\ref{eq:lda_phi_equation}) and Eq. (\ref{eq:lda_gamma_equation}).
\UNTIL $\gamma_{dk}$ converged.
\ENDFOR
\STATE \emph{M-step:}
\STATE Set $\lambda_{kw}(t+1)$ from Eq. (\ref{eq:lda_lambda_equation_online}).
\UNTIL $\mathcal{L}$ converged.
\end{algorithmic}
\end{algorithm}

\section{Description of ToT\label{sec:Description-of-ToT}}

\begin{figure}
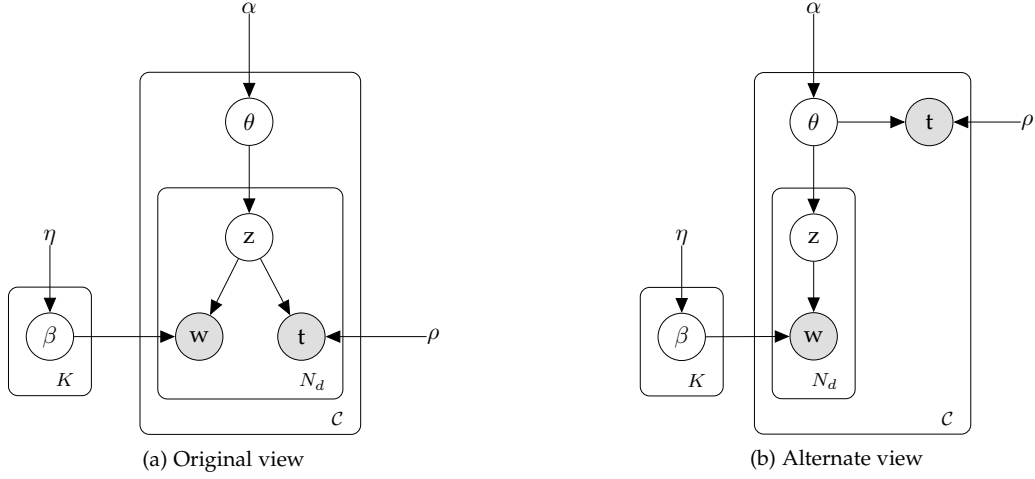
   
\centering   
\resizebox{.9\linewidth}{!}{
\begin{minipage}{.5\textwidth}
\centering
\tikz{ %
		\node[const] (alpha) {$\alpha$} ; %
		\node[latent, below=1.25cm of alpha] (theta) {$\theta$} ; %
		\node[latent, below=of theta] (z) {z} ; %
		\node[const, left=2.5cm of z] (eta) {$\eta$} ; %
		\node[latent, below=1cm of eta] (beta) {$\beta$} ; %
		\node[obs, right=1.5cm of beta] (w) {w} ; %

		\node[obs, right=0.8cm of w] (t) {t} ; %
		\node[const, right=1.5cm of t] (rho) {$\rho$} ; %

		\plate[inner sep=0.25cm, yshift=0.12cm] {plate0} {(beta)} {$K$}; %
		\plate[inner sep=0.25cm, yshift=0.12cm] {plate1} {(z) (w) (t)} {$N_d$}; %
		\plate[inner sep=0.25cm, yshift=0.12cm] {plate2} {(theta) (plate1)} {$\mathcal{C}$}; %
		\edge {alpha} {theta} ; %
		\edge {theta} {z} ; %
		\edge {z,beta} {w} ; %
		\edge {eta} {beta} ; %
		\edge {rho} {t} ; %
		\edge {z} {t} ; %
	} 

\subcaption{Original view}   
\label{fig:dag_tot} 
\end{minipage}%
\begin{minipage}{.5\textwidth}
\centering
\tikz{ %
		\node[const] (alpha) {$\alpha$} ; %
		\node[latent, below=1.25cm of alpha] (theta) {$\theta$} ; %
		\node[obs, right=of theta] (t) {t} ; %
		\node[const, right=of t] (rho) {$\rho$} ; %
		\node[latent, below=of theta] (z) {z} ; %
		\node[const, left=1.5cm of z] (eta) {$\eta$} ; %
  		\node[latent, below=1cm of eta] (beta) {$\beta$} ; %
    	\node[obs, below=0.75cm of z] (w) {w} ; %

		\plate[inner sep=0.25cm, yshift=0.12cm] {plate0} {(beta)} {$K$}; %
		\plate[inner sep=0.25cm, yshift=0.12cm] {plate1} {(z) (w)} {$N_d$}; %
		\plate[inner sep=0.25cm, yshift=0.12cm] {plate2} {(theta) (t) (plate1)} {$\mathcal{C}$}; %
		\edge {alpha} {theta} ; %
		\edge {theta} {z} ; %
		\edge {theta} {t} ; %
		\edge {z,beta} {w} ; %
		\edge {rho} {t} ; %
		\edge {eta} {beta} ; %
	} 

\subcaption{Alternate view}   
\label{fig:dag_tot_alternate} 
\end{minipage}%
}
\caption{Directed acyclic graphs for Topics over Time}
\end{figure}

The main references for the results in this appendix are \citet{Wang2006,Fang2017}.
The generative process for ToT can be described as:
\begin{enumerate}
\item Draw $K$ multinomial parameters $\boldsymbol{\beta_{k}}$ from $K$ Dirichlet distributions
with parameters $\boldsymbol{\eta_{k}}$.
\item For each document $d$, draw a set of multinomial parameters $\boldsymbol{\theta_{d}}$ from a symmetric
Dirichlet distribution with parameters $\boldsymbol{\alpha}$. Then, for each word
index $i\in\{1,\ldots, N_{d}\}$ in document $d$:
\begin{enumerate}
\item Draw a topic $z_{di}$ from a multinomial with parameters $\boldsymbol{\theta_{d}}$;
\item Draw a word $w_{di}$ from a multinomial with parameters $\boldsymbol{\beta_{z_{di}}}$;
\item Draw a timestamp $t_{di}$ from a Beta distribution with parameters $\boldsymbol{\rho}_{z_{di}}$.
\end{enumerate}
\end{enumerate}
See Fig. \ref{fig:dag_tot} for the directed acyclic graph representation
of this generative process. The corresponding log-likelihood has the
form:

\begin{equation}
\begin{aligned}\mathcal{L}^{\text{ToT}}=\mathcal{L}^{\text{LDA}}+ & \sum_{d=1}^{D}\sum_{i=1}^{N_{d}}\log\underbrace{p\left(t_{di}|\boldsymbol{\rho}_{z_{di}}\right)}_{\text{Beta}}.\end{aligned}
\label{eq:TopicsOverTimeLogLikehood}
\end{equation}
where $\mathcal{L}^{\text{LDA}}$ is given by Eq. (\ref{eq:LDAlogLikehood}). 

Gibbs sampling was originally proposed as an approximate inference
method in \citet{Wang2006}. However, we will follow the variational
inference method as described in \citet{Fang2017} since it more closely
resembles our approach. The proposed evidence lower bound (ELBO) is 

\begin{equation}
L^{\text{ToT-ELBO}}=\left\langle \mathcal{L^{\text{ToT}}}-\log q^{\text{LDA}}\right\rangle _{q^{\text{LDA}}},
\end{equation}
where $q^{\text{LDA}}$ is given by Eq. (\ref{eq:LDAVariationalAnsatz}).
Notice that no variational distribution has been introduced to approximate
the timestamp distribution, so the $\boldsymbol{\rho}_{k}$ parameters
will be determined exactly (i.e. the variational ansatz $q$ is the
same as in LDA, $q^{\text{LDA}}$). The maximization of this ELBO
leads to the following equations

\begin{equation}
\phi_{dwk}\propto\exp\left[\left\langle \log\theta_{dk}\right\rangle _{q^{\text{LDA}}}+\left\langle \log\beta_{kw}\right\rangle _{q^{\text{LDA}}}+\left(\rho_{k}^{1}-1\right)\log t_{d}+\left(\rho_{k}^{2}-1\right)\log\left[1-t_{d}\right]-\log B\left(\boldsymbol{\rho}_{k}\right)\right],\label{eq:tot_phi_equation}
\end{equation}

\begin{equation}
\frac{\sum_{dw}n_{dw}\phi_{dwk}\text{\ensuremath{\log}}t_{d}}{\sum_{dw}n_{dw}\phi_{dwk}}=\Psi\left(\rho_{k}^{1}\right)-\Psi\left(\rho_{k}^{1}+\rho_{k}^{2}\right),\label{eq:tot_rho1_equation}
\end{equation}

\begin{equation}
\frac{\sum_{dw}n_{dw}\phi_{dwk}\text{\ensuremath{\log}}\left(1-t_{d}\right)}{\sum_{dw}n_{dw}\phi_{dwk}}=\Psi\left(\rho_{k}^{2}\right)-\Psi\left(\rho_{k}^{1}+\rho_{k}^{2}\right).\label{eq:tot_rho2_equation}
\end{equation}
where $\Psi\left(x\right):=d\log\Gamma\left(x\right)/dx$ is the digamma
function. The Dirichlet averages appearing in Eq. (\ref{eq:tot_phi_equation})
are given by Eqs. (\ref{eq:theta_q_mean}) - (\ref{beta_q_mean}), and the
update equations for $\lambda_{kw}$ and $\gamma_{dk}$ are exactly
the same as in standard LDA (see Eqs. (\ref{eq:lda_gamma_equation}) - (\ref{eq:lda_lambda_equation})). The update equations for $\boldsymbol{\rho}_{k}$
cannot be solved directly due to the appearance of the digamma function
in Eqs. (\ref{eq:tot_rho1_equation}) - (\ref{eq:tot_rho2_equation}), so numerical
algorithms must be used. 

The general optimization algorithm is summarized in Algorithm \ref{alg:BatchToT}. Note that no online optimization version was proposed in \citet{Wang2006, Fang2017}. This was expected, since Eqs. (\ref{eq:tot_rho1_equation}) - (\ref{eq:tot_rho2_equation}) would yield unstable mini-batch updates.

\begin{algorithm}
\caption{\label{alg:BatchToT} Batch variational ToT}
\begin{algorithmic}
\STATE Initialize variational parameters $\gamma_{dk}$, $\lambda_{kw}$ and $\boldsymbol{\rho}_k$.
\REPEAT
\STATE \emph{E-step:}
\FOR{$d=1$ to $D$}
\REPEAT
\STATE Set $\phi_{dwk}$, $\gamma_{dk}$ from Eq. (\ref{eq:tot_phi_equation}) and Eq. (\ref{eq:lda_gamma_equation}).
\UNTIL $\gamma_{dk}$ converged.
\ENDFOR
\UNTIL $\mathcal{L}$ converged.
\STATE \emph{M-step:}
\STATE Set $\lambda_{kw}$, $\boldsymbol{\rho}_k$ from Eqs. (\ref{eq:lda_lambda_equation}), (\ref{eq:tot_rho1_equation}) and (\ref{eq:tot_rho2_equation}).
\end{algorithmic}
\end{algorithm}

\section{Integrability analysis of the Beta-prior distribution}\label{sec:integrability-analysis}

Consider the normalization function $f(\nu,\boldsymbol{\chi})$ in the definition
of the Beta-prior distribution,
\begin{equation}
f\left(\nu,\boldsymbol{\chi}\right)^{-1}:=\int_{0}^{\infty}\int_{0}^{\infty}d^{2}\boldsymbol{\rho}\exp\left[\nu\left(\boldsymbol{\rho}\cdot\boldsymbol{\chi}-\log B\left(\boldsymbol{\rho}\right)\right)\right]. \label{eq:normalization}
\end{equation}
Its expression in polar coordinates is
\begin{equation}
I_{g()}\left(\nu,\boldsymbol{\chi}\right)=\int_{0}^{\pi/2}d\theta\int_{0}^{\infty} g\left(\boldsymbol{\rho}\right)\exp\left[\nu\left(\rho\left( \chi^1\cos{\theta}+\chi^2\sin{\theta}\right) - \log B\left(\rho\cos{\theta},\rho\sin{\theta}\right) \right)\right] \rho \,d\rho.\label{eq:bpn_pol}
\end{equation}
We will study the large $\rho$ behavior of this integral. In order for Eq. (\ref{eq:bpn_pol}) to be finite, the exponential term must satisfy:
\begin{equation}
\nu \rho \underbrace{\left[\cos{\theta}\left(\chi^1 - \log{\frac{\cos{\theta}}{\cos{\theta} + \sin{\theta}}}\right) + \sin{\theta}\left(\chi^2 - \log{\frac{\sin{\theta}}{\cos{\theta} + \sin{\theta}}}\right)\right]}_{F\left(\theta, \boldsymbol{\chi}\right)}<0, \label{eq:integrability_condition}
\end{equation}
where we have made use of the Stirling approximation to write the leading part of the logarithm term in Eq. (\ref{eq:bpn_pol}) as
\begin{equation}
\log B\left(\rho\cos{\theta},\rho\sin{\theta}\right) \simeq  \rho\left(\cos{\theta}\log{\frac{\cos{\theta}}{\cos{\theta} + \sin{\theta}}} + \sin{\theta}\log{\frac{\sin{\theta}}{\cos{\theta} + \sin{\theta}}}\right). \label{eq:large_rho_log}
\end{equation}
Let us denote $\theta'$ the angle for which $F\left(\theta, \boldsymbol{\chi}\right)$ takes its maximum value, i.e.
\begin{equation}
\frac{\partial F\left(\theta', \boldsymbol{\chi}\right)}{\partial \theta} = -\chi^1\sin{\theta'} + \chi^2\cos{\theta'} - \left(\cos{\theta'}\log{\frac{\cos{\theta'}}{\cos{\theta'} + \sin{\theta'}}} - \sin{\theta'}\log{\frac{\sin{\theta'}}{\cos{\theta'} + \sin{\theta'}}}\right) = 0. \label{eq:max_F_condition}
\end{equation}
From Eq. (\ref{eq:max_F_condition}) and the definition of $F\left(\theta, \boldsymbol{\chi}\right)$ in Eq. (\ref{eq:integrability_condition}), it follows that
\begin{equation}
e^{\chi^1} + e^{\chi^2} = e^{\cos{\theta'}F\left(\theta'\right)} \frac{\cos{\theta'}}{\cos{\theta'}+\sin{\theta'}} + \frac{\sin{\theta'}}{\cos{\theta'+\sin{\theta'}}}.
\end{equation}
Therefore,
\begin{equation}
e^{\chi^1} + e^{\chi^2} < 1 \iff F\left(\theta', \boldsymbol{\chi}\right)<0.
\end{equation}

\section{Asymptotic expansion of Beta-prior integrals\label{sec:Asymptotic-expansion-of}}

The normalization function $f(\nu,\boldsymbol{\chi})$ in the definition
of the Beta-prior distribution, Eq. (\ref{eq:normalization}), cannot be calculated analytically. However, asymptotic methods can
be applied to obtain an approximation. To do so, consider the large
$\nu\gg1$ expansion of the various moments of the Beta-prior distribution,
\begin{equation}
I_{g()}\left(\nu,\boldsymbol{\chi}\right):=\int_{0}^{\infty}\int_{0}^{\infty}d^{2}\boldsymbol{\rho}\,g\left(\boldsymbol{\rho}\right)\exp\left[\nu\left(\boldsymbol{\rho}\cdot\boldsymbol{\chi}-\log B\left(\boldsymbol{\rho}\right)\right)\right],\label{eq:IModelIntegral}
\end{equation}
where $g\left(\boldsymbol{\rho}\right)$ is an arbitrary function
of $\boldsymbol{\rho}$. By the Laplace method \citep{Gelman2013},
we obtain that
\begin{equation}
I_{g()}\left(\nu,\boldsymbol{\chi}\right)\simeq e^{\nu\omega\left(\boldsymbol{\rho}_{0}\right)}\frac{2\pi}{\nu\sqrt{\left|\det\left(\omega_{ij}\right)\right|}}\left(g\left(\boldsymbol{\rho}_{0}\right)+O\left(\nu^{-1}\right)\right),\label{eq:asymptotic_expansion}
\end{equation}
where we have defined
\begin{equation}
\omega\left(\boldsymbol{\rho}\right)\coloneqq\boldsymbol{\rho}\cdot\text{\ensuremath{\boldsymbol{\chi}}}-\log B\left(\boldsymbol{\rho}\right),
\end{equation}
\begin{equation}
\boldsymbol{\rho}_{0}\coloneqq\mathrm{argmax_{\boldsymbol{\rho}}\left\{ \omega\left(\boldsymbol{\rho}\right)\right\} },
\end{equation}
\begin{equation}
\omega_{ij}\coloneqq\frac{\partial^{2}\omega\left(\boldsymbol{\rho}_{0}\right)}{\partial\rho^{i}\partial\rho^{j}}.
\end{equation}
Notice that the asymptotic expansion of the normalization integral
$f\left(\nu,\boldsymbol{\chi}\right)^{-1}$ can be obtained from Eq. (\ref{eq:asymptotic_expansion})
by setting $g(\boldsymbol{\rho})=1$. From this, the moments under
the Beta-prior in Eq. (\ref{eq:phiUpdate}) can now be estimated asymptotically
as:
\begin{alignat}{1}
\left\langle \rho_{i}\right\rangle  & =\frac{1}{\nu}\frac{\partial}{\partial\chi_{i}}\log f\left(\nu,\boldsymbol{\chi}\right)^{-1}\nonumber \\
 & \simeq\rho_{0i}+O\left(\nu^{-1}\right)\label{eq:AsymptoticMomentRho}
\end{alignat}
\begin{align}
\left\langle \log B\left(\boldsymbol{\rho}\right)\right\rangle  & =-\frac{\partial}{\partial\nu}\log f\left(\nu,\boldsymbol{\chi}\right)^{-1}+\boldsymbol{\chi}\cdot\left\langle \boldsymbol{\rho}\right\rangle \nonumber \\
 & \simeq-\omega\left(\boldsymbol{\rho}_{0}\right)+O\left(\nu^{-1}\right)+\boldsymbol{\chi}\cdot\left\langle \boldsymbol{\rho}\right\rangle \nonumber \\
 & =\log B\left(\boldsymbol{\rho}_{0}\right)+O\left(\nu^{-1}\right)\label{eq:AsymptoticMomentLogB}
\end{align}

\section{Balancing hyper-parameter in ToT and BToT \label{sec:Balancing-hyper-parameter}}

Both the BToT model and the original ToT need some adjustments to balance
the relative influence of words and timestamps in the log-likelihood,
as described in \citet{Wang2006,Fang2017}. Notice that there is a
single timestamp per document, but in Fig. \ref{fig:dag_btot}
and \ref{fig:dag_tot} its probability is repeated as many times
as words are in the document. To deal with this issue, both references
propose to change the log-likelihood in Eq. (\ref{eq:BToT_ts_logLikehood})
by adding a ``balancing hyperparameter'' $0<\delta<1$. This new
quasi-likelihood\footnote{The term quasi-likelihood is used with a different meaning in other
contexts.} now reads
\begin{align}
\mathcal{L^{\delta}}_{ts} & =\delta\left[\sum_{d=1}^{D}\sum_{i=1}^{N_{d}}\left\{ \left(\rho_{z_{di}}^{1}-1\right)\log t_{di}+\left(\rho_{z_{di}}^{2}-1\right)\log\left[1-t_{di}\right]-\log B\left(\boldsymbol{\rho}_{z_{di}}\right)\right\} \right]\nonumber \\
 & +\nu\sum_{k=1}^{K}\left\{ \boldsymbol{\rho}_{k}\cdot\boldsymbol{\chi}-\log B\left(\boldsymbol{\rho}_{k}\right)\right\} +K\log f\left(\nu,\boldsymbol{\chi}\right)
\end{align}
Normalization is lost, but for the purpose of making inferences with
the posterior, one can use this quasi-likelihood as an ordinary one
because only quotients are involved (normalization can be recovered in a Gibbs sampling approach \citealt{Wang2006}).
The same variational treatment described in Section \ref{subsubsec:btot-vational-inference}
can be used, and results in the following modifications: 
\begin{itemize}
\item The variational Beta-prior updates, Eq. (\ref{eq:variationalBetaPrior}),
are changed to
\begin{equation}
\mu_{k}=\nu+\delta\,N_{k}
\end{equation}
\begin{equation}
\boldsymbol{\psi}_{k}=\mu_{k}^{-1}\left(\nu\boldsymbol{\chi}+\delta\,N_{k}\boldsymbol{l}_{k}\right).
\end{equation}
\item The multinomial updates, Eq. (\ref{eq:phiUpdate}), are changed to
\begin{align}
\phi_{dwk} & \propto\exp\left[\left\langle \log\beta_{kw}\right\rangle _{q^{\text{LDA}}}+\left\langle \log\theta_{dk}\right\rangle _{q^{\text{LDA}}}\right.\nonumber \\
 & \left.+\delta\left\{ \left(\left\langle \rho_{k}^{1}\right\rangle _{q_{k}\left(\boldsymbol{\rho}_{k}\right)}-1\right)\log t_{d}+\left(\left\langle \rho_{k}^{2}\right\rangle _{q_{k}\left(\boldsymbol{\rho}_{k}\right)}-1\right)\log\left[1-t_{d}\right]-\left\langle \log B\left(\boldsymbol{\rho}_{k}\right)\right\rangle _{q_{k}\left(\boldsymbol{\rho}_{k}\right)}\right\} \right]\label{eq:phiUpdate-1}
\end{align}
\item The online updates, Eqs. (\ref{eq:btot_mu_online_update}) - (\ref{eq:btot_psi_online_update}),
now read
\begin{equation}
\mu'_{k}\left(t+1\right)=\left(1-\rho_{t}\right)\mu'_{k}\left(t\right)+\rho_{t}\left(\nu+\frac{D}{S}\sum_{d\in MB}\delta\,N_{dk}\right)
\end{equation}
\begin{equation}
\boldsymbol{\psi}'_{k}\left(t+1\right)=\left(1-\rho_{t}\right)\boldsymbol{\psi}'_{k}\left(t\right)+\rho_{t}\left(\nu\boldsymbol{\chi}+\frac{D}{S}\sum_{d\in MB}\delta\,N_{dk}\boldsymbol{lt}_{d}\right).
\end{equation}
\end{itemize}
Notice that, in most of these modifications, one simply changes $N_{dk}\rightarrow\delta\,N_{dk}.$

It is not difficult to show that one can absorb the hyperparameter
$\delta$ into a re-scaling of the Beta-prior parameters $\boldsymbol{\chi}\rightarrow\boldsymbol{\chi}/\delta$
(if one is not too picky with respect to the domain of integration
of $d^{2}\boldsymbol{\rho}$), so in order to impose effectively this
hyperparameter it is necessary to strongly adjust the priors. Therefore,
this model would not be appropriate if one wants to stick to a Bayesian
approach.

In \citet{Wang2006} it is suggested that a natural value for $\delta$
would be the inverse of the average number of words per document in
the collection, where it is also discussed a connection to a similar
model in which timestamps are generated just once per document (see
Fig. \ref{fig:dag_tot_alternate}). However, this model severely
underestimates the influence of the time dimension. 

\bibliography{library}

\begin{thebibliography}{38}
\expandafter\ifx\csname natexlab\endcsname\relax\def\natexlab#1{#1}\fi
\providecommand{\url}[1]{\texttt{#1}}
\providecommand{\href}[2]{#2}
\providecommand{\path}[1]{#1}
\providecommand{\DOIprefix}{doi:}
\providecommand{\ArXivprefix}{arXiv:}
\providecommand{\URLprefix}{URL: }
\providecommand{\Pubmedprefix}{pmid:}
\providecommand{\doi}[1]{\href{http://dx.doi.org/#1}{\path{#1}}}
\providecommand{\Pubmed}[1]{\href{pmid:#1}{\path{#1}}}
\providecommand{\bibinfo}[2]{#2}
\ifx\xfnm\relax \def\xfnm[#1]{\unskip,\space#1}\fi
\bibitem[{Al-Ghossein et~al.(2018)Al-Ghossein, Murena, Abdessalem, Barr{\'{e}} and Cornu{\'{e}}jols}]{Al-Ghossein2018}
\bibinfo{author}{Al-Ghossein, M.}, \bibinfo{author}{Murena, P.A.}, \bibinfo{author}{Abdessalem, T.}, \bibinfo{author}{Barr{\'{e}}, A.}, \bibinfo{author}{Cornu{\'{e}}jols, A.}, \bibinfo{year}{2018}.
\newblock \bibinfo{title}{{Adaptive Collaborative Topic Modeling for Online Recommendation}}, in: \bibinfo{booktitle}{Proceedings of the 12th ACM Conference on Recommender Systems}, \bibinfo{publisher}{Association for Computing Machinery}, \bibinfo{address}{Vancouver, British Columbia, Canada}. pp. \bibinfo{pages}{338--346}.
\newblock \DOIprefix\doi{10.1145/3240323.3240363}.
\bibitem[{Asghari et~al.(2020)Asghari, Sierra-Sosa and Elmaghraby}]{Asghari2020}
\bibinfo{author}{Asghari, M.}, \bibinfo{author}{Sierra-Sosa, D.}, \bibinfo{author}{Elmaghraby, A.S.}, \bibinfo{year}{2020}.
\newblock \bibinfo{title}{{A topic modeling framework for spatio-temporal information management}}.
\newblock \bibinfo{journal}{Information Processing \& Management} \bibinfo{volume}{57}, \bibinfo{pages}{102340}.
\newblock \URLprefix \url{https://doi.org/10.1016/j.ipm.2020.102340}, \DOIprefix\doi{10.1016/j.ipm.2020.102340}.
\bibitem[{Banda et~al.(2021)Banda, Tekumalla, Wang, Yu, Liu, Ding, Artemova, Tutubalina and Chowell}]{Banda2021}
\bibinfo{author}{Banda, J.M.}, \bibinfo{author}{Tekumalla, R.}, \bibinfo{author}{Wang, G.}, \bibinfo{author}{Yu, J.}, \bibinfo{author}{Liu, T.}, \bibinfo{author}{Ding, Y.}, \bibinfo{author}{Artemova, E.}, \bibinfo{author}{Tutubalina, E.}, \bibinfo{author}{Chowell, G.}, \bibinfo{year}{2021}.
\newblock \bibinfo{title}{{A Large-Scale COVID-19 Twitter Chatter Dataset for Open Scientific Research - An International Collaboration}}.
\newblock \bibinfo{journal}{Epidemiologia} \bibinfo{volume}{2}, \bibinfo{pages}{315--324}.
\newblock \DOIprefix\doi{10.3390/epidemiologia2030024}.
\bibitem[{Bird et~al.(2009)Bird, Klein and Loper}]{nltk_book}
\bibinfo{author}{Bird, S.}, \bibinfo{author}{Klein, E.}, \bibinfo{author}{Loper, E.}, \bibinfo{year}{2009}.
\newblock \bibinfo{title}{{Natural language processing with Python: analyzing text with the natural language toolkit}}.
\newblock \bibinfo{publisher}{O'Reilly Media, Inc.}
\bibitem[{Bishop(2006)}]{Bishop2007}
\bibinfo{author}{Bishop, C.M.}, \bibinfo{year}{2006}.
\newblock \bibinfo{title}{{Pattern Recognition and Machine Learning (Information Science and Statistics)}}.
\newblock \bibinfo{edition}{1} ed., \bibinfo{publisher}{Springer}, \bibinfo{address}{New York, USA}.
\bibitem[{Blei and Lafferty(2006)}]{Blei2006}
\bibinfo{author}{Blei, D.M.}, \bibinfo{author}{Lafferty, J.D.}, \bibinfo{year}{2006}.
\newblock \bibinfo{title}{{Dynamic Topic Models}}, in: \bibinfo{booktitle}{Proceedings of the 23rd International Conference on Machine Learning}, \bibinfo{publisher}{Association for Computing Machinery}, \bibinfo{address}{Pittsburgh, Pennsylvania, USA}. pp. \bibinfo{pages}{113--120}.
\newblock \DOIprefix\doi{10.1145/1143844.1143859}.
\bibitem[{Blei et~al.(2003)Blei, Ng and Jordan}]{Blei2003}
\bibinfo{author}{Blei, D.M.}, \bibinfo{author}{Ng, A.Y.}, \bibinfo{author}{Jordan, M.I.}, \bibinfo{year}{2003}.
\newblock \bibinfo{title}{{Latent Dirichlet Allocation}}.
\newblock \bibinfo{journal}{Journal of Machine Learning Research} \bibinfo{volume}{3}, \bibinfo{pages}{993--1022}.
\newblock \DOIprefix\doi{10.5555/944919.944937}.
\bibitem[{Bunk and Krestel(2018)}]{Bunk2018}
\bibinfo{author}{Bunk, S.}, \bibinfo{author}{Krestel, R.}, \bibinfo{year}{2018}.
\newblock \bibinfo{title}{{WELDA: Enhancing Topic Models by Incorporating Local Word Context}}, in: \bibinfo{booktitle}{Proceedings of the ACM/IEEE Joint Conference on Digital Libraries}, pp. \bibinfo{pages}{293--302}.
\newblock \DOIprefix\doi{10.1145/3197026.3197043}.
\bibitem[{Burel et~al.(2021)Burel, Farrell and Alani}]{Burel2021}
\bibinfo{author}{Burel, G.}, \bibinfo{author}{Farrell, T.}, \bibinfo{author}{Alani, H.}, \bibinfo{year}{2021}.
\newblock \bibinfo{title}{{Demographics and topics impact on the co-spread of COVID-19 misinformation and fact-checks on Twitter}}.
\newblock \bibinfo{journal}{Information Processing \& Management} \bibinfo{volume}{58}, \bibinfo{pages}{102732}.
\newblock \URLprefix \url{https://doi.org/10.1016/j.ipm.2021.102732}, \DOIprefix\doi{10.1016/j.ipm.2021.102732}.
\bibitem[{Churchill and Singh(2021)}]{Churchill2021}
\bibinfo{author}{Churchill, R.}, \bibinfo{author}{Singh, L.}, \bibinfo{year}{2021}.
\newblock \bibinfo{title}{{The Evolution of Topic Modeling}}.
\newblock \bibinfo{journal}{ACM Computing Surveys} \DOIprefix\doi{10.1145/3507900}.
\bibitem[{Das et~al.(2018)Das, Mueller, Hargrove, Roman and Baden}]{Das2018}
\bibinfo{author}{Das, A.}, \bibinfo{author}{Mueller, F.}, \bibinfo{author}{Hargrove, P.}, \bibinfo{author}{Roman, E.}, \bibinfo{author}{Baden, S.}, \bibinfo{year}{2018}.
\newblock \bibinfo{title}{{Doomsday: Predicting Which Node Will Fail When on Supercomputers}}, in: \bibinfo{booktitle}{SC18: International Conference for High Performance Computing, Networking, Storage and Analysis}, \bibinfo{publisher}{IEEE}. pp. \bibinfo{pages}{108--121}.
\newblock \DOIprefix\doi{10.1109/SC.2018.00012}.
\bibitem[{Devlin et~al.(2019)Devlin, Chang, Lee and Toutanova}]{Devlin2019}
\bibinfo{author}{Devlin, J.}, \bibinfo{author}{Chang, M.W.}, \bibinfo{author}{Lee, K.}, \bibinfo{author}{Toutanova, K.}, \bibinfo{year}{2019}.
\newblock \bibinfo{title}{{BERT: Pre-training of deep bidirectional transformers for language understanding}}.
\newblock \bibinfo{journal}{NAACL HLT 2019 - 2019 Conference of the North American Chapter of the Association for Computational Linguistics: Human Language Technologies - Proceedings of the Conference} \bibinfo{volume}{1}, \bibinfo{pages}{4171--4186}.
\newblock \href{http://arxiv.org/abs/1810.04805}{\tt arXiv:1810.04805}.
\bibitem[{Dieng et~al.(2020)Dieng, Ruiz and Blei}]{Dieng2020}
\bibinfo{author}{Dieng, A.B.}, \bibinfo{author}{Ruiz, F.J.}, \bibinfo{author}{Blei, D.M.}, \bibinfo{year}{2020}.
\newblock \bibinfo{title}{{Topic modeling in embedding spaces}}.
\newblock \bibinfo{journal}{Transactions of the Association for Computational Linguistics} \bibinfo{volume}{8}, \bibinfo{pages}{439--453}.
\newblock \DOIprefix\doi{10.1162/tacl_a_00325}.
\bibitem[{Fang et~al.(2017)Fang, Macdonald, Ounis, Habel and Yang}]{Fang2017}
\bibinfo{author}{Fang, A.}, \bibinfo{author}{Macdonald, C.}, \bibinfo{author}{Ounis, I.}, \bibinfo{author}{Habel, P.}, \bibinfo{author}{Yang, X.}, \bibinfo{year}{2017}.
\newblock \bibinfo{title}{{Exploring Time-Sensitive Variational Bayesian Inference LDA for Social Media Data}}, in: \bibinfo{booktitle}{Advances in Information Retrieval. ECIR 2017. Lecture Notes in Computer Science}, \bibinfo{publisher}{Springer International Publishing}, \bibinfo{address}{Cham}. pp. \bibinfo{pages}{252--265}.
\newblock \DOIprefix\doi{10.1007/978-3-319-56608-5_20}.
\bibitem[{Gelman et~al.(2013)Gelman, Carlin, Stern, Dunson, Vehtari and Rubin}]{Gelman2013}
\bibinfo{author}{Gelman, A.}, \bibinfo{author}{Carlin, J.B.}, \bibinfo{author}{Stern, H.S.}, \bibinfo{author}{Dunson, D.B.}, \bibinfo{author}{Vehtari, A.}, \bibinfo{author}{Rubin, D.B.}, \bibinfo{year}{2013}.
\newblock \bibinfo{title}{{Bayesian Data Analysis}}.
\newblock \bibinfo{edition}{3rd ed.} ed., \bibinfo{publisher}{Chapman and Hall/CRC}, \bibinfo{address}{New York}.
\newblock \DOIprefix\doi{10.1201/b16018}.
\bibitem[{Grootendorst(2022)}]{Grootendorst2022}
\bibinfo{author}{Grootendorst, M.}, \bibinfo{year}{2022}.
\newblock \bibinfo{title}{{BERTopic: Neural topic modeling with a class-based TF-IDF procedure}}.
\newblock \URLprefix \url{http://arxiv.org/abs/2203.05794}, \href{http://arxiv.org/abs/2203.05794}{\tt arXiv:2203.05794}. \bibinfo{note}{arXiv preprint}.
\bibitem[{Hoffman et~al.(2010)Hoffman, Bach and Blei}]{Hoffman2010}
\bibinfo{author}{Hoffman, M.D.}, \bibinfo{author}{Bach, F.}, \bibinfo{author}{Blei, D.M.}, \bibinfo{year}{2010}.
\newblock \bibinfo{title}{{Online Learning for Latent Dirichlet Allocation}}, in: \bibinfo{booktitle}{Advances in Neural Information Processing Systems}, \bibinfo{publisher}{Curran Associates, Inc.}. p. \bibinfo{pages}{856–864}.
\newblock \DOIprefix\doi{10.5555/2997189.2997285}.
\bibitem[{Hoffman et~al.(2013)Hoffman, Blei, Wang and Paisley}]{Hoffman2013}
\bibinfo{author}{Hoffman, M.D.}, \bibinfo{author}{Blei, D.M.}, \bibinfo{author}{Wang, C.}, \bibinfo{author}{Paisley, J.}, \bibinfo{year}{2013}.
\newblock \bibinfo{title}{{Stochastic Variational Inference}}.
\newblock \bibinfo{journal}{Journal of Machine Learning Research} \bibinfo{volume}{14}, \bibinfo{pages}{1303--1347}.
\bibitem[{Hofmann(1999)}]{Hofmann1999}
\bibinfo{author}{Hofmann, T.}, \bibinfo{year}{1999}.
\newblock \bibinfo{title}{{Probabilistic Latent Semantic Analysis}}, in: \bibinfo{booktitle}{Proceedings of the Fifteenth Conference on Uncertainty in Artificial Intelligence}, \bibinfo{publisher}{Morgan Kaufmann Publishers Inc.}, \bibinfo{address}{Stockholm, Sweden}. pp. \bibinfo{pages}{289----296}.
\newblock \DOIprefix\doi{10.5555/2073796.2073829}.
\bibitem[{Jelodar et~al.(2019)Jelodar, Wang, Yuan, Feng, Jiang, Li and Zhao}]{Jelodar2019}
\bibinfo{author}{Jelodar, H.}, \bibinfo{author}{Wang, Y.}, \bibinfo{author}{Yuan, C.}, \bibinfo{author}{Feng, X.}, \bibinfo{author}{Jiang, X.}, \bibinfo{author}{Li, Y.}, \bibinfo{author}{Zhao, L.}, \bibinfo{year}{2019}.
\newblock \bibinfo{title}{{Latent Dirichlet allocation (LDA) and topic modeling: models, applications, a survey}}.
\newblock \bibinfo{journal}{Multimedia Tools and Applications} \bibinfo{volume}{78}, \bibinfo{pages}{15169--15211}.
\newblock \DOIprefix\doi{10.1007/s11042-018-6894-4}.
\bibitem[{Lau and Lau(1991)}]{Lau1991}
\bibinfo{author}{Lau, H.S.}, \bibinfo{author}{Lau, A.H.L.}, \bibinfo{year}{1991}.
\newblock \bibinfo{title}{{Effective procedures for estimating beta distribution's parameters and their confidence intervals}}.
\newblock \bibinfo{journal}{Journal of Statistical Computation and Simulation} \bibinfo{volume}{38}, \bibinfo{pages}{139--150}.
\newblock \DOIprefix\doi{10.1080/00949659108811325}.
\bibitem[{Li et~al.(2019a)Li, Xie, Jiang, Zhou and Huang}]{Li2019}
\bibinfo{author}{Li, X.}, \bibinfo{author}{Xie, Q.}, \bibinfo{author}{Jiang, J.}, \bibinfo{author}{Zhou, Y.}, \bibinfo{author}{Huang, L.}, \bibinfo{year}{2019}a.
\newblock \bibinfo{title}{{Identifying and monitoring the development trends of emerging technologies using patent analysis and Twitter data mining: The case of perovskite solar cell technology}}.
\newblock \bibinfo{journal}{Technological Forecasting and Social Change} \bibinfo{volume}{146}, \bibinfo{pages}{687--705}.
\newblock \DOIprefix\doi{10.1016/j.techfore.2018.06.004}.
\bibitem[{Li et~al.(2019b)Li, Zhang and Ouyang}]{Li2019a}
\bibinfo{author}{Li, X.}, \bibinfo{author}{Zhang, J.}, \bibinfo{author}{Ouyang, J.}, \bibinfo{year}{2019}b.
\newblock \bibinfo{title}{{Dirichlet multinomial mixture with variational manifold regularization: Topic modeling over short texts}}.
\newblock \bibinfo{journal}{Proceedings of the AAAI Conference on Artificial Intelligence} \bibinfo{volume}{33}, \bibinfo{pages}{7884--7891}.
\newblock \DOIprefix\doi{10.1609/aaai.v33i01.33017884}.
\bibitem[{McInnes and Healy(2017)}]{McInnes2017}
\bibinfo{author}{McInnes, L.}, \bibinfo{author}{Healy, J.}, \bibinfo{year}{2017}.
\newblock \bibinfo{title}{{Accelerated Hierarchical Density Based Clustering}}.
\newblock \bibinfo{journal}{IEEE International Conference on Data Mining Workshops, ICDMW} \bibinfo{volume}{2017-Novem}, \bibinfo{pages}{33--42}.
\newblock \DOIprefix\doi{10.1109/ICDMW.2017.12}, \href{http://arxiv.org/abs/arXiv:1705.07321v2}{\tt arXiv:arXiv:1705.07321v2}.
\bibitem[{McInnes et~al.(2018)McInnes, Healy and Melville}]{McInnes2018}
\bibinfo{author}{McInnes, L.}, \bibinfo{author}{Healy, J.}, \bibinfo{author}{Melville, J.}, \bibinfo{year}{2018}.
\newblock \bibinfo{title}{{UMAP: Uniform Manifold Approximation and Projection for Dimension Reduction}}.
\newblock \URLprefix \url{http://arxiv.org/abs/1802.03426}, \href{http://arxiv.org/abs/1802.03426}{\tt arXiv:1802.03426}. \bibinfo{note}{arXiv preprint}.
\bibitem[{Mele et~al.(2019)Mele, Bahrainian and Crestani}]{Mele2019}
\bibinfo{author}{Mele, I.}, \bibinfo{author}{Bahrainian, S.A.}, \bibinfo{author}{Crestani, F.}, \bibinfo{year}{2019}.
\newblock \bibinfo{title}{{Event mining and timeliness analysis from heterogeneous news streams}}.
\newblock \bibinfo{journal}{Information Processing \& Management} \bibinfo{volume}{56}, \bibinfo{pages}{969--993}.
\newblock \URLprefix \url{https://doi.org/10.1016/j.ipm.2019.02.003}, \DOIprefix\doi{10.1016/j.ipm.2019.02.003}.
\bibitem[{Poumay and Ittoo(2021)}]{Poumay2021}
\bibinfo{author}{Poumay, J.}, \bibinfo{author}{Ittoo, A.}, \bibinfo{year}{2021}.
\newblock \bibinfo{title}{{HTMOT : Hierarchical Topic Modelling Over Time}}.
\newblock \bibinfo{journal}{arXiv preprint} \href{http://arxiv.org/abs/2112.03104}{\tt arXiv:2112.03104}.
\bibitem[{Qi et~al.(2020)Qi, Zhang, Zhang, Bolton and Manning}]{Qi2020}
\bibinfo{author}{Qi, P.}, \bibinfo{author}{Zhang, Y.}, \bibinfo{author}{Zhang, Y.}, \bibinfo{author}{Bolton, J.}, \bibinfo{author}{Manning, C.D.}, \bibinfo{year}{2020}.
\newblock \bibinfo{title}{{Stanza: A Python Natural Language Processing Toolkit for Many Human Languages}}, in: \bibinfo{booktitle}{Proceedings of the 58th Annual Meeting of the Association for Computational Linguistics: System Demonstrations}, \bibinfo{publisher}{Association for Computational Linguistics}, \bibinfo{address}{Online}. pp. \bibinfo{pages}{101--108}.
\newblock \DOIprefix\doi{10.18653/v1/2020.acl-demos.14}.
\bibitem[{Reimers and Gurevych(2019)}]{Reimers2019}
\bibinfo{author}{Reimers, N.}, \bibinfo{author}{Gurevych, I.}, \bibinfo{year}{2019}.
\newblock \bibinfo{title}{{Sentence-BERT: Sentence embeddings using siamese BERT-networks}}.
\newblock \bibinfo{journal}{EMNLP-IJCNLP 2019 - 2019 Conference on Empirical Methods in Natural Language Processing and 9th International Joint Conference on Natural Language Processing, Proceedings of the Conference} , \bibinfo{pages}{3982--3992}\DOIprefix\doi{10.18653/v1/d19-1410}, \href{http://arxiv.org/abs/1908.10084}{\tt arXiv:1908.10084}.
\bibitem[{R{\"{o}}der et~al.(2015)R{\"{o}}der, Both and Hinneburg}]{Roder2015}
\bibinfo{author}{R{\"{o}}der, M.}, \bibinfo{author}{Both, A.}, \bibinfo{author}{Hinneburg, A.}, \bibinfo{year}{2015}.
\newblock \bibinfo{title}{{Exploring the Space of Topic Coherence Measures}}, in: \bibinfo{booktitle}{Proceedings of the Eighth ACM International Conference on Web Search and Data Mining}, \bibinfo{publisher}{Association for Computing Machinery}, \bibinfo{address}{Shanghai, China}. pp. \bibinfo{pages}{399--408}.
\newblock \DOIprefix\doi{10.1145/2684822.2685324}.
\bibitem[{Shi et~al.(2018)Shi, Bryan, Bhamidipati, Zhao, Zhang and Ma}]{Shi2018}
\bibinfo{author}{Shi, Y.}, \bibinfo{author}{Bryan, C.}, \bibinfo{author}{Bhamidipati, S.}, \bibinfo{author}{Zhao, Y.}, \bibinfo{author}{Zhang, Y.}, \bibinfo{author}{Ma, K.L.}, \bibinfo{year}{2018}.
\newblock \bibinfo{title}{{MeetingVis: Visual Narratives to Assist in Recalling Meeting Context and Content}}.
\newblock \bibinfo{journal}{IEEE Transactions on Visualization and Computer Graphics} \bibinfo{volume}{24}, \bibinfo{pages}{1918--1929}.
\newblock \DOIprefix\doi{10.1109/TVCG.2018.2816203}.
\bibitem[{Vaswani et~al.(2017)Vaswani, Shazeer, Parmar, Uszkoreit, Jones, Gomez, Kaiser and Polosukhin}]{Vaswani2017}
\bibinfo{author}{Vaswani, A.}, \bibinfo{author}{Shazeer, N.}, \bibinfo{author}{Parmar, N.}, \bibinfo{author}{Uszkoreit, J.}, \bibinfo{author}{Jones, L.}, \bibinfo{author}{Gomez, A.N.}, \bibinfo{author}{Kaiser, {\L}.}, \bibinfo{author}{Polosukhin, I.}, \bibinfo{year}{2017}.
\newblock \bibinfo{title}{{Attention Is All You Need}}, in: \bibinfo{booktitle}{31st Conference on Neural Information Processing Systems}, p. \bibinfo{pages}{6000–6010}.
\newblock \DOIprefix\doi{10.1109/2943.974352}.
\bibitem[{Viegas et~al.(2020)Viegas, Cunha, Gomes, Pereira, Rocha and Goncalves}]{Viegas2020}
\bibinfo{author}{Viegas, F.}, \bibinfo{author}{Cunha, W.}, \bibinfo{author}{Gomes, C.}, \bibinfo{author}{Pereira, A.}, \bibinfo{author}{Rocha, L.}, \bibinfo{author}{Goncalves, M.}, \bibinfo{year}{2020}.
\newblock \bibinfo{title}{{CluHTM - Semantic Hierarchical Topic Modeling based on CluWords}}, in: \bibinfo{booktitle}{Proceedings of the 58th Annual Meeting of the Association for Computational Linguistics}, \bibinfo{publisher}{Association for Computational Linguistics}, \bibinfo{address}{Online}. pp. \bibinfo{pages}{8138--8150}.
\newblock \DOIprefix\doi{10.18653/v1/2020.acl-main.724}.
\bibitem[{Wang et~al.(2008)Wang, Blei and Heckerman}]{Wang2008}
\bibinfo{author}{Wang, C.}, \bibinfo{author}{Blei, D.}, \bibinfo{author}{Heckerman, D.}, \bibinfo{year}{2008}.
\newblock \bibinfo{title}{{Continuous Time Dynamic Topic Models}}, in: \bibinfo{booktitle}{Proceedings of the Twenty-Fourth Conference on Uncertainty in Artificial Intelligence}, \bibinfo{publisher}{AUAI Press}, \bibinfo{address}{Helsinki, Finland}. pp. \bibinfo{pages}{579--586}.
\newblock \DOIprefix\doi{10.5555/3023476.3023545}.
\bibitem[{Wang and McCallum(2006)}]{Wang2006}
\bibinfo{author}{Wang, X.}, \bibinfo{author}{McCallum, A.}, \bibinfo{year}{2006}.
\newblock \bibinfo{title}{{Topics over Time: A Non-Markov Continuous-Time Model of Topical Trends}}, in: \bibinfo{booktitle}{Proceedings of the 12th ACM SIGKDD International Conference on Knowledge Discovery and Data Mining}, \bibinfo{publisher}{Association for Computing Machinery}, \bibinfo{address}{Philadelphia, PA, USA}. pp. \bibinfo{pages}{424--433}.
\newblock \DOIprefix\doi{10.1145/1150402.1150450}.
\bibitem[{Xu et~al.(2014)Xu, Shi, Qiao, Zhu, Jung, Lee and Choi}]{Xu2014}
\bibinfo{author}{Xu, S.}, \bibinfo{author}{Shi, Q.}, \bibinfo{author}{Qiao, X.}, \bibinfo{author}{Zhu, L.}, \bibinfo{author}{Jung, H.}, \bibinfo{author}{Lee, S.}, \bibinfo{author}{Choi, S.P.}, \bibinfo{year}{2014}.
\newblock \bibinfo{title}{{Author-Topic over Time (AToT): A Dynamic Users' Interest Model}}, in: \bibinfo{booktitle}{Mobile, ubiquitous, and intelligent computing}, \bibinfo{publisher}{Springer Berlin Heidelberg}, \bibinfo{address}{Berlin, Heidelberg}. pp. \bibinfo{pages}{239--245}.
\newblock \DOIprefix\doi{10.1007/978-3-642-40675-1_37}.
\bibitem[{Xue et~al.(2021)Xue, {Qiping Shen}, Li, Han and Chu}]{Xue2021}
\bibinfo{author}{Xue, J.}, \bibinfo{author}{{Qiping Shen}, G.}, \bibinfo{author}{Li, Y.}, \bibinfo{author}{Han, S.}, \bibinfo{author}{Chu, X.}, \bibinfo{year}{2021}.
\newblock \bibinfo{title}{{Dynamic Analysis on Public Concerns in Hong Kong-Zhuhai-Macao Bridge: Integrated Topic and Sentiment Modeling Approach}}.
\newblock \bibinfo{journal}{Journal of Construction Engineering and Management} \bibinfo{volume}{147}, \bibinfo{pages}{04021049}.
\newblock \DOIprefix\doi{10.1061/(ASCE)CO.1943-7862.0002066}.
\bibitem[{Xue et~al.(2020)Xue, Shen, Li, Wang and Zafar}]{Xue2020}
\bibinfo{author}{Xue, J.}, \bibinfo{author}{Shen, G.Q.}, \bibinfo{author}{Li, Y.}, \bibinfo{author}{Wang, J.}, \bibinfo{author}{Zafar, I.}, \bibinfo{year}{2020}.
\newblock \bibinfo{title}{{Dynamic Stakeholder-Associated Topic Modeling on Public Concerns in Megainfrastructure Projects: Case of Hong Kong-Zhuhai-Macao Bridge}}.
\newblock \bibinfo{journal}{Journal of Management in Engineering} \bibinfo{volume}{36}, \bibinfo{pages}{04020078}.
\newblock \DOIprefix\doi{10.1061/(asce)me.1943-5479.0000845}.

\end{thebibliography}

\end{document}